\documentclass[11pt, a4paper, logo]{gdmstyle/googledeepmind}

\pdftrailerid{redacted}

\makeatletter
\renewcommand\bibentry[1]{\nocite{#1}{\frenchspacing\@nameuse{BR@r@#1\@extra@b@citeb}}}
\makeatother

\usepackage{algorithm}
\usepackage{algpseudocode}
\newcommand{\algrule}[1][.2pt]

\algnewcommand\algorithmicinput{\textbf{Input:}}
\algnewcommand\Input{\item[\algorithmicinput]}
\algnewcommand\algorithmicoutput{\textbf{Output:}}
\algnewcommand\Output{\item[\algorithmicoutput]}
\algnewcommand\algorithmicempty{~}
\algnewcommand\Empty{\item[\algorithmicempty]}

\usepackage{mathtools}

\usepackage{kantlipsum, lipsum}
\usepackage{dsfont}
\usepackage{gdmstyle/gdm-colors}
\usepackage{bm}      %

\usepackage{ifthen}
\newboolean{isanon}
\setboolean{isanon}{false}

\newboolean{islong}
\setboolean{islong}{true}
\usepackage{xcolor}
\usepackage{xurl}
\usepackage{graphicx}
\usepackage{amsmath}

\usepackage{pifont}%
\usepackage[sort,numbers,square]{natbib}
\usepackage[pagebackref=true,colorlinks=true,allcolors=colorgreenfull]{hyperref}
\usepackage{cleveref}
\usepackage{scalerel}
\usepackage{caption}
\usepackage{wrapfig}
\usepackage{subcaption}
\usepackage{bbding}
\usepackage{placeins}
\usepackage{microtype}
\usepackage{csquotes}
\usepackage{enumitem}
\usepackage{xspace}
\usepackage{booktabs}
\usepackage{mathtools}
\usepackage{amsthm}

\usepackage{tikz}
\usetikzlibrary{calc}
\usetikzlibrary{angles,quotes} %

\usepackage[super]{nth}
\usepackage{multirow}
\usepackage{rotating}
\usepackage{amssymb}
\usepackage[export]{adjustbox}
\usepackage{subcaption}
\usepackage{tcolorbox}
\everypar{\looseness=-1}

\usetikzlibrary{decorations.markings}
\usetikzlibrary{shapes.geometric, positioning, arrows, shapes.symbols, calc}

\renewcommand*{\backrefalt}[4]{%
    \ifcase #1 \footnotesize{(Not cited.)}%
    \or        \footnotesize{(p.~#2)}%
    \else      \footnotesize{(pp.~#2)}%
    \fi}
\newtheorem{observation}{Observation}

\Crefname{observation}{Observation}{Observations}
\Crefname{hypothesis}{Hypothesis}{Hypotheses}
\newtheorem{remark}{Remark}
\newtheorem{assumption}{Assumption}
\newtheorem*{assumption*}{Assumption}
\Crefname{assumption}{Assumption}{Assumptions}
\newtheorem{lemma}{Lemma}
\newtheorem*{lemma*}{Lemma}
\Crefname{lemma}{Lemma}{Lemmas}

\Crefname{remark}{Remark}{Remarks}
\Crefname{proposition}{Proposition}{Propositions}

\newcommand*\X{\mathcal{X}}

\newcommand*{\eg}{e.g.,\@\xspace}
\newcommand*{\versus}{vs.\@\xspace}

\newcommand*{\ie}{i.e.,\@\xspace}

\newcommand*{\wrt}{w.r.t.\@\xspace}

\let\originalleft\left
\let\originalright\right
\renewcommand{\left}{\mathopen{}\mathclose\bgroup\originalleft}
\renewcommand{\right}{\aftergroup\egroup\originalright}

\def\eqref#1{eq.~\ref{#1}}

\def\X{\mathcal{X}}

\def\vx{{\bm{x}}}
\def\vy{{\bm{y}}}

\DeclareMathAlphabet{\mathsfit}{\encodingdefault}{\sfdefault}{m}{sl}
\SetMathAlphabet{\mathsfit}{bold}{\encodingdefault}{\sfdefault}{bx}{n}

\newcommand{\KL}{$\mathrm{KL}$\@\xspace}
\newcommand{\Kl}{\mathrm{KL}}
\newcommand{\Esp}{\mathbb{E}}

\DeclareMathOperator*{\argmax}{argmax}

\newcommand{\TODO}[2]{}

\newcommand{\WA}{WA\@\xspace}
\newcommand{\SFT}{SFT\@\xspace}
\newcommand{\RLHF}{RLHF\@\xspace}

\newcommand{\WARM}{\emph{WARM}\@\xspace}
\newcommand{\WARP}{\emph{WARP}\@\xspace}
\newcommand{\EMA}{\emph{EMA}\@\xspace}
\newcommand{\SLERP}{\emph{SLERP}\@\xspace}

\newcommand{\LERP}{\emph{LERP}\@\xspace}
\newcommand{\LITI}{\emph{LITI}\@\xspace}
\newcommand{\GEMMA}{Gemma\@\xspace}
\newcommand{\GEMMAB}{Gemma "7B"\@\xspace}
\newcommand{\WISE}{WiSE-FT\@\xspace}

\title{\WARP: On the Benefits of Weight Averaged Rewarded Policies}

\correspondingauthor{alexandrerame@google.com}

\keywords{Alignment, RLHF, LLM, Model Merging}

\author{Alexandre~Ramé}
\author{Johan~Ferret}
\author{Nino~Vieillard}
\author{Robert~Dadashi}
\author{Léonard~Hussenot}
\author{Pierre-Louis~Cedoz}
\author{Pier~Giuseppe~Sessa}
\author{Sertan~Girgin}
\author{Arthur~Douillard}
\author{Olivier~Bachem}
\affil{Google DeepMind}

\begin{abstract}
	Reinforcement learning from human feedback (\RLHF) aligns large language models (LLMs) by encouraging their generations to have high rewards, using a reward model trained on human preferences.
To prevent the forgetting of pre-trained knowledge, \RLHF usually incorporates a \KL regularization; this forces the policy to remain close to its supervised fine-tuned initialization, though it hinders the reward optimization.
To tackle the trade-off between \KL and reward, in this paper we introduce a novel alignment strategy named Weight Averaged Rewarded Policies (\WARP).
\WARP merges policies in the weight space at three distinct stages.
First, it uses the exponential moving average of the policy as a dynamic anchor in the \KL regularization.
Second, it applies spherical interpolation to merge independently fine-tuned policies into a new enhanced one.
Third, it linearly interpolates between this merged model and the initialization, to recover features from pre-training.
This procedure is then applied iteratively, with each iteration's final model used as an advanced initialization for the next, progressively refining the \KL-reward Pareto front, achieving superior rewards at fixed \KL.
Experiments with \GEMMA policies validate that \WARP improves their quality and alignment, outperforming other open-source LLMs.

\end{abstract}

\begin{document}

\maketitle

\section{Introduction}
\label{sec:introduction}
\textbf{LLM alignment.}
Conversational agents like Gemini \cite{gemini2023,reid2024gemini} and GPT-4 \cite{openai2023gpt4}, along with their open-weight counterparts like \GEMMA \cite{team2024gemma}, have demonstrated remarkable abilities in complex tasks including mathematics, coding, and tool use \cite{bubeck2023sparks}. These capabilities largely emerge from pre-training on next-token prediction \cite{radford2018gen,radford2019language}, subsequently refined through supervised fine-tuning (\SFT) \cite{JMLR:v21:20-074,wei2022finetuned}.
As these LLMs become more powerful, aligning them with human values becomes increasingly crucial to ensure safe deployment~\cite{amodei2016concrete,hendrycks2022x}.
To this end, reinforcement learning from human feedback (\RLHF) has become the prominent strategy \cite{christiano2017deep, ziegler2019fine, stiennon2020learning}, first learning a reward model (RM) on human preferences, before optimizing the LLM to maximize predicted~rewards.

\textbf{Challenges in \RLHF.}
However, \RLHF introduces several unresolved challenges \cite{casper2023open}.
First, the limited scope of fine-tuning, often restricted to relatively small datasets, can lead to excessive specialization and catastrophic forgetting~\cite{french1992semi} of the broad and diverse knowledge acquired during pre-training \cite{goodfellow2013empirical, li2017learning, kirkpatrick2017overcoming, kumar2022finetuning}. Such \emph{alignment tax} \cite{ouyang2022training} can degrade the LLM's reasoning capabilities and performance on NLP benchmarks \cite{dong2023abilities,lin2024mitigating}.
Second, maximizing an imperfect RM presents several issues on its own, as the LLM can learn to exploit loopholes in the RM \cite{faultyreward2016,pan2022the} when it deviates significantly from its initialization \cite{gao2022scaling}. Such \emph{reward hacking} \cite{askell2021general, skalse2022defining} can produce outputs that are linguistically flawed \cite{lewis2017deal}, excessively verbose \cite{singhal2023long}, or sycophantic \cite{perez2022discovering, sharma2023towards}, thereby raising misalignment \cite{taylor2016alignment, ngo2022alignment} and safety \cite{amodei2016concrete, hendrycks2022x} concerns.
Finally, \RLHF can reduce the diversity of generations \cite{kirk2024understanding}, potentially leading to policy collapse \cite{Moalla2024NoRN,hamilton2024detecting}. Such \emph{loss of diversity} limits use in creative or exploratory tasks and can result in the LLM systematically refusing to answer.
Overall, achieving high rewards based on an imperfect RM on a selected distribution of prompts is insufficient due to potential reward misspecification and distribution shifts upon deployment.

\begin{figure*}[t!]
	\begin{center}
		\begin{subfigure}[b]{0.63\textwidth}
			\includegraphics[width=\textwidth]{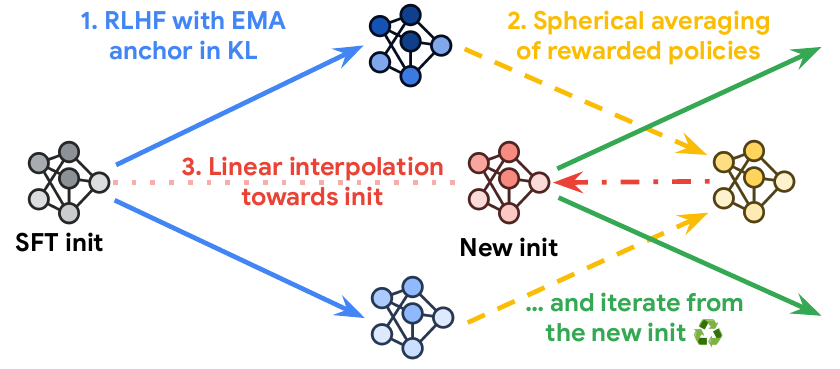}
			\caption{\WARP with three model merging stages, applicable iteratively.}%
			\label{fig:main:warp}
		\end{subfigure}%
		\hfill%
		\begin{subfigure}[b]{0.35\textwidth}%
			\includegraphics[width=\textwidth]{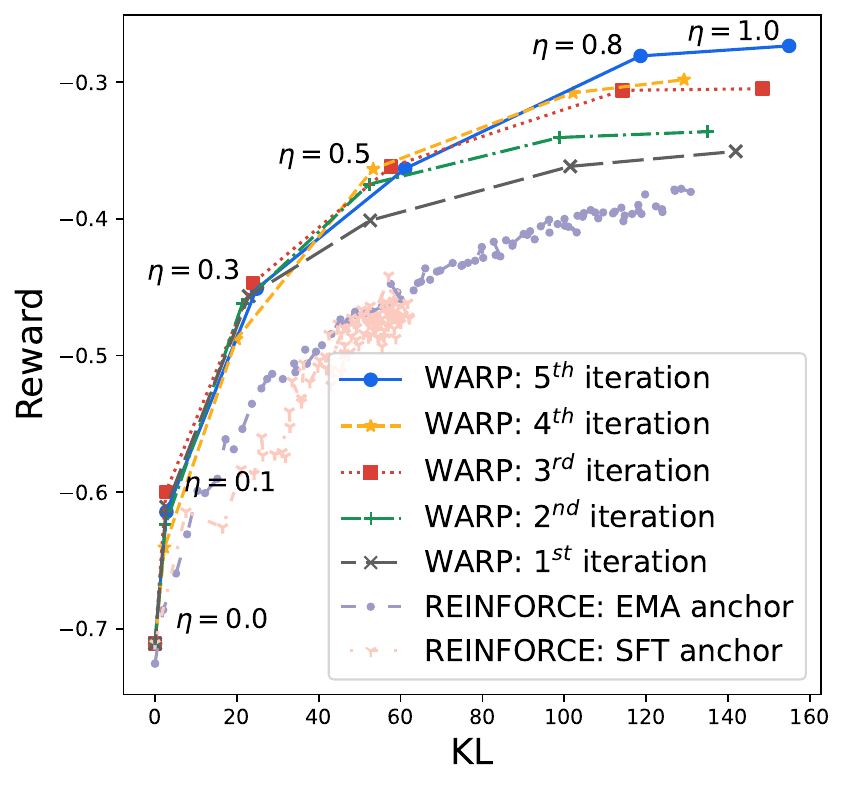}
			\caption{\KL-reward Pareto front.}%
			\label{fig:main:paretofront}%
		\end{subfigure}
	\end{center}%
	\caption{%
		\Cref{fig:main:warp} illustrates the \RLHF alignment process with \WARP from a supervised fine-tuned (\SFT) LLM. \WARP uses model merging by weight averaging at three different stages.
		First, the exponential moving average (\EMA)~\cite{izmailov2018} of the policy serves as the anchor for \KL regularization~\cite{jaques2017sequence}.
		Second, the independently fine-tuned policies are merged by spherical linear interpolation (\SLERP)~\cite{shoemake1985animating} of their task vectors~\cite{2022arXiv221204089I}.
		Third, we interpolate towards the initialization (\LITI)~\cite{Wortsman2022robust}, revealing a Pareto front of solutions as we slide the interpolating coefficient $\eta$ from $1$ to $0$. This results in the \enquote{\WARP: \nth{1} iteration} curve from \Cref{fig:main:paretofront} which improves over the REINFORCE~\cite{williams1992simple} fine-tuning trajectories. Critically, iteratively using a point from this Pareto front as an advanced initialization for the next episode \WARP improves performance. Details in \Cref{fig:warp_vs_iterations}.}%
	\label{fig:main}%
\end{figure*}

\textbf{RL with \KL regularization.}
To address these issues, previous works constrained the reward optimization by integrating a Kullback-Leibler (\KL) regularization \cite{jaques2017sequence,geist2019theory}, using the \SFT initialization as the anchor.
As clarified in \Cref{sec:context}, this \KL regularization forces the policy to remain close to its initialization \cite{lazaridou2020multi,lu2020countering}, mitigating forgetting and reward hacking \cite{gao2022scaling}. However, employing the \SFT model as the anchor may lead to reward underfitting: indeed, there is a fundamental tension between reducing \KL and maximizing reward. Thus, different policies should be compared in terms of \KL-reward Pareto optimality as in \Cref{fig:main:paretofront}, where the $x$-axis is the \KL and the $y$-axis is the reward as estimated by the RM, with the optimal policies located in the top-left of the plot.

\textbf{On model merging by weight averaging.}
To improve the trade-off between \KL and reward during \RLHF,
we leverage the ability to merge LLMs by weight averaging (\WA) \cite{utans1996weight}.
\WA relies on the linear mode connectivity ~\cite{Frankle2020,Neyshabur2020}, an empirical observation revealing linear paths of high performance between models fine-tuned from a shared pre-trained initialization.
Model merging was shown to improve robustness under distribution shifts \cite{izmailov2018, Wortsman2022ModelSA, rame2022diwa} by promoting generalization and reducing memorization \cite{rame2024warm}, to combine models' abilities \cite{2022arXiv221204089I,ilharco2022patching,rame2023rewarded}, to reduce forgetting in continual learning \cite{stojanovski2022momentum}, to enable collaborative \cite{updatablemachinelearning} and distributed \cite{douillard2023diloco} learning at scale, without computational overheads at inference time.
Model merging is increasingly adopted within the open-source community \cite{goddard2024arcee,nathanlambert2024modelmerging}, leading to state-of-the-art models in specialized domains \cite{labrak2024biomistral} but also significant advancements on general-purpose benchmarks \cite{neuralbeagle24, maximelabonne2024}. 
In particular, while \WA was initially mostly used for discriminative tasks \cite{Wortsman2022ModelSA} such as reward modeling \cite{rame2024warm}, it is now becoming popular for generative tasks \cite{rofin2022linear,akiba2024evolutionary}; its use in \KL-constrained \RLHF has already shown preliminary successes in a few recent works \cite{rame2023rewarded,noukhovitch2023language,lin2024mitigating,liu2024decoding,gorbatovski2024learn,munos2023nash}, further elaborated in \Cref{sec:related}.

\textbf{\WARP.}
In this paper, we propose Weight Averaged Rewarded Policies (\WARP), a simple strategy for aligning LLMs, illustrated in \Cref{fig:main:warp} and detailed in \Cref{sec:model}. \WARP is designed to optimize the \KL-reward Pareto front of solutions, as demonstrated in \Cref{fig:main:paretofront}.
\WARP uses three variants of \WA at three different stages of the alignment procedure, for three distinct reasons.

\begin{tcolorbox}[colback=colorblue,
		colframe=black,
		arc=4pt,
		boxsep=0.3pt,
	]%
	\textbf{Stage 1: \emph{Exponential Moving Average} (\EMA).} During RL fine-tuning, instead of regularizing the policy towards the \SFT initialization, \WARP uses the policy's own exponential moving average \cite{polyak1992acceleration} as a dynamic updatable anchor in the \KL. This stage enables stable exploration with distillation from a mean teacher \cite{tarvainen2017mean} and annealed constraint.
\end{tcolorbox}%
\begin{tcolorbox}[colback=coloryellow,
		colframe=black,
		arc=4pt,
		boxsep=0.3pt,
	]
	\textbf{Stage 2: \emph{Spherical Linear intERPolation of task vectors} (\SLERP).}
	Considering $M$ policies RL fine-tuned independently with their own \EMA anchor, we merge them by spherical linear interpolation \cite{shoemake1985animating} of their task vectors \cite{2022arXiv221204089I}. This stage creates a merged model with higher reward by combining the strengths of the $M$ individual policies.
\end{tcolorbox}%
\begin{tcolorbox}[colback=colorred,
		colframe=black,
		arc=4pt,
		boxsep=0.3pt,
	]
	\textbf{Stage 3: \emph{Linear Interpolation Towards Initialization} (\LITI).}
	Considering the merged policy from \SLERP, \WARP linearly interpolates towards the initialization, akin to \WISE \cite{Wortsman2022robust}.
	This stage allows to run through an improved Pareto-front simply by adjusting the interpolating coefficient $\eta$ between $1$ (high reward but high \KL) and $0$ (small \KL but small reward). Critically, selecting an intermediate value for $0<\eta<1$ offers a balanced model that can serve as a new, improved initialization for subsequent iterations of \WARP.
\end{tcolorbox}%

\textbf{Experiments and discussion.} In \Cref{sec:expe}, we validate the efficacy of \WARP for the fine-tuning of \GEMMAB \cite{team2024gemma}. Finally, in \Cref{sec:discussion}, we discuss the connections between \WARP, the distributed learning literature \cite{updatablemachinelearning, douillard2023diloco} and iterated amplification \cite{christiano2018supervising}, illustrating how \WARP embodies their principles to enable scaling post-training, for continuous alignment and improvement of LLMs.

\section{Context and notations}
\label{sec:context}%
\textbf{RL for LLMs.}
We consider a transformer \cite{vaswani2017transformer} LLM $f\left(\cdot, \theta\right)$ parameterized by $\theta$.
Following the foundation model paradigm \cite{bommasani2021opportunities} and the principles of transfer learning \cite{oquab2014learning}, those weights are trained via a three-stage procedure: pre-training through next token prediction, supervised fine-tuning resulting in $\theta_\mathrm{sft}$, and ultimately, RLHF \cite{christiano2017deep,ouyang2022training} to optimize a reward $r$ as determined by a RM trained to reflect human preferences.
In this RL stage, $\theta$ defines a policy $\pi_{\theta}(\cdot \mid \vx)$ by auto-regressively generating token sequences $\vy$ from the prompt $\vx$.
The primary objective is to find weights maximizing the average reward over a dataset of prompts $\X$: $\argmax_\theta \Esp_{\vx \in \X} \Esp_{\vy \sim \pi_{\theta}(\cdot \mid \vx)} \Big[ r \left(\vx, \vy \right) \Big]$.

\textbf{\KL \versus reward.}
Optimizing solely for $r$ can (i) forget general abilities from pre-training \cite{french1992semi} as an alignment tax \cite{ouyang2022training, lin2024mitigating}, (ii) hack the reward \cite{askell2021general,skalse2022defining} leading to potential misalignment, or (iii) reduce the diversity of possible generations \cite{kirk2024understanding} (confirmed in \Cref{app:dlp}).
To mitigate these risks, a \KL regularization is usually integrated to balance fidelity to the initialization and high rewards:%
\begin{equation}%
	\argmax_\theta \Esp_{\vx \in \X} \left[ \Esp_{\vy \sim \pi_{\theta}(\cdot \mid \vx)}  r \left(\vx, \vy \right) - \beta \Kl\left(\pi_\theta(\cdot\mid\vx) \| \pi_{\theta_\mathrm{anchor}}(\cdot\mid\vx)\right)\right],%
	\label{eq:argmaxrkl}%
\end{equation}%
where $\theta_\mathrm{anchor}\leftarrow\theta_\mathrm{sft}$ and $\beta$ is the regularization strength, with high values leading to low \KL though also lower reward.
The reward function adjusted with this \KL is $r(\vx, \vy)-\beta \log \left(\frac{\pi_\theta(\vy \mid \vx)}{\pi_{\theta_\mathrm{anchor}}(\vy \mid \vx)}\right).$
Our base RL algorithm is a variant of REINFORCE~\cite{williams1992simple}. This choices follows recent \RLHF works \cite{roit2023factually,lee2023rlaif,rame2024warm} and the findings from \cite{li2023remax,Tajwar2024ppodpo,ahmadian2024back} that, in terms of \KL-reward Pareto optimality, REINFORCE performs better than the more complex PPO~\cite{schulman2017proximal} and also better than various offline algorithms such as DPO~\cite{rafailov2023direct}, IPO \cite{azar2023general} or RAFT \cite{dong2023raft}.
Practitioners then typically employ early stopping to select an optimal point on the training trajectory based on their specific~use cases.%

\section{\WARP}
\label{sec:model}
We introduce a novel alignment strategy named Weight Averaged Rewarded Policies (\WARP), illustrated in \Cref{fig:main:warp} and described in \Cref{alg:warp} below.
\WARP merges LLMs in the weight space to enhance the \KL-reward Pareto front of policies.
The following \Cref{sec:model:ema,sec:model:slerp,sec:model:liti} describe the motivations behind applying three distinct variants of \WA at the three different stages of \WARP.
In particular, we summarize the key insights as observations, that will be experimentally validated in \Cref{sec:expe} (and in \Cref{app:slerp:empirical,app:expe}), and theoretically motivated in \Cref{app:slerp:theoryslerpvslerp} when possible.
Overall, \WARP outperforms other RL alignment strategies, without any memory or inference overhead at test time.
However, training \WARP is costly, requiring multiple RL runs at each iteration: see \Cref{sec:discussion} for a detailed discussion on the required compute scaling.
\begin{center}%
\begin{minipage}{1.0\textwidth}%
	\begin{algorithm}[H]%
		\caption{\WARP for \KL-reward Pareto optimal alignment}%
		\label{alg:warp}%
		\begin{algorithmic}[1]%
			\Input {Weights $\theta_\mathrm{sft}$ pre-trained and supervised fine-tuned
				\Statex{\hspace{1.5em} Reward model $r$, prompt dataset $\X$, optimizer $\mathrm{Opt}$}
				\Statex{\hspace{1.5em} $I$ iterations with $M$ RL runs each for $T$ training steps}
				\Statex{\hspace{1.5em} $\mu$ \EMA update rate, $\eta$ \LITI update rate}
				\algrule}
			\State Define $\theta_\mathrm{init} \leftarrow \theta_\mathrm{sft}$
			\For{iteration $i$ from $1$ to $I$}
			\For{run $m$ from $1$ to $M$} \Comment{Run in parallel}
			\State Define $\theta^m, \theta_\mathrm{ema}^m\leftarrow\theta_\mathrm{init}$
			\For{step $t$ from $1$ to $T$}
			\State Generate completion $\vy\sim\pi_{\theta^m}(\cdot \mid \vx)$ for $\vx\in\X$
			\State Compute $r_{\beta}(\vy)\leftarrow r \left(\vx, \vy \right) - \beta \log \frac{\pi_{\theta^m}(\vy \mid \vx)}{\pi_{\theta_\mathrm{ema}^m}(\vy \mid \vx)}$ \Comment{\KL regularized reward}
			\State Update $\theta^m \leftarrow \mathrm{Opt}\left(\theta^m, r_{\beta}(\vy) \nabla_{\theta} \left[\log \pi_{\theta^m}\left(\vy \mid \vx\right)\right]\right)$ \Comment{Policy gradient}
			\State Update $\theta_\mathrm{ema}^m \leftarrow (1-\mu) \cdot \theta_\mathrm{ema}^m + \mu \cdot \theta^m$ \Comment{\Cref{eq:ema}: update anchor}
			\EndFor
			\EndFor
			\State Define $\theta_\mathrm{slerp}^{i}\leftarrow \operatorname{slerp}\left(\theta_\mathrm{init}, \{\theta^m\}_{m=1}^{M}, \lambda=\frac{1}{M}\right)$ \Comment{\Cref{eq:slerp}: merge $M$ weights}
			\State Update $\theta_\mathrm{init} \leftarrow (1-\eta) \cdot \theta_\mathrm{init} + \eta \cdot \theta_\mathrm{slerp}^{i}$ \Comment{\Cref{eq:liti}: interpolate towards init}
			\EndFor
			\algrule
			\Output \KL-reward Pareto front of weights $\{(1-\eta) \cdot \theta_\mathrm{sft} + \eta \cdot \theta_\mathrm{slerp}^{I}\mid0\leq\eta\leq1\}$
		\end{algorithmic}
	\end{algorithm}
\end{minipage}
\end{center}

\subsection{\textcolor{colorbluefull}{Stage 1: exponential moving average as a dynamic anchor in \KL regularization}}
\label{sec:model:ema}
\textbf{\EMA anchor.}
\KL-regularized methods typically use the \SFT initialization as a static anchor~\cite{jaques2017sequence,roit2023factually}, but in RL for control tasks, it is common to regularly update the anchor~\cite{schulman2015trust, abdolmaleki2018maximum}. In this spirit, \WARP uses the policy's own exponential moving average (\EMA)~\cite{polyak1992acceleration}, updated throughout the RL fine-tuning process such as, at each training step with $\mu=0.01$:
\begin{equation}
	\theta_\mathrm{ema} \leftarrow (1-\mu) \cdot \theta_\mathrm{ema} + \mu \cdot \theta_\mathrm{policy}.
	\label{eq:ema}
	\tag{\EMA}
\end{equation}
Using $\theta_\mathrm{ema}$ as the anchor $\theta_\mathrm{anchor}$ in \Cref{eq:argmaxrkl} provides several benefits, outlined below.
\begin{observation}[\EMA]
	Policies trained with an exponential moving average anchor benefit from automatic annealing of the \KL regularization and from distillation from a dynamic mean teacher~\cite{tarvainen2017mean}.
	Empirical evidence in \Cref{sec:expe:ema}.%
	\label{obs:ema}%
\end{observation}
\textbf{Benefits from \EMA.}
Unlike a static SFT anchor, the dynamic nature of an \EMA anchor induces a gradual automatic annealing and relaxation of the \KL regularization. Specifically, the policy is initially strongly tied to the \SFT initialization, and then progressively unleashed, allowing for more aggressive gradient updates later in training, leading to higher rewards. Moreover, by progressively incorporating knowledge from the training, \EMA acts as slow weight \cite{stojanovski2022momentum,lee2024grokfast}, and thus performing better than the initialization. But, by also maintaining essential information from the initialization, \EMA can even perform better than the final policy's weights;  
studies \cite{szegedy2016rethinking, izmailov2018, arpit2021ensemble} (see \cite{moralesbrotons2024exponential} for a review), and specifically \cite{kaddour2022stop} within the context of LLMs, indicate that averaging checkpoints over steps improves internal representations and thus predictions. Then, \EMA guides the policy by \KL distillation \cite{hinton2015distilling} of high-quality target predictions, akin to a mean teacher~\cite{tarvainen2017mean} for self-supervised \cite{sohn20_fixmat,he2020momentum, oquab2024dinov, Caron_2021_ICCV, grill2020bootstrap} learning.
This also relates to deep RL techniques where \EMA stabilizes exploration toward a Nash equilibrium \cite{awheda2013exponential, awheda2016exponential, gorbatovski2024learn, munos2023nash}, and approximates mirror descent \cite{bubeck2015convex,geist2019theory,tomar2020mirror}.

\subsection{\textcolor{coloryellowfull}{Stage 2: spherical linear interpolation of independently rewarded policies}}
\label{sec:model:slerp}
\textbf{\SLERP.}
While \EMA helps for a single RL and a fixed compute budget, it faces limitations due to the similarity of the weights collected along a single fine-tuning \cite{rame2022diwa}.
In this second stage, we merge $M$ weights RL fine-tuned independently (each with their own \EMA anchor). This follows model soups from \citet{Wortsman2022ModelSA} and its variants \cite{rame2022diwa,rame2022recycling} showing that \WA improves generalization, and that task vectors~\cite{2022arXiv221204089I} (the difference between fine-tuned weights and their initialization) can be arithmetically manipulated by linear interpolation (\LERP)~\cite{utans1996weight}. Yet, this time, we use spherical linear interpolation (\SLERP) \cite{shoemake1985animating}, illustrated in \Cref{fig:merge} and defined below for $M=2$:
\begin{equation}
	\operatorname{slerp}\left(\theta_\mathrm{init}, \theta^1, \theta^2, \lambda\right)=\theta_\mathrm{init} + \frac{\sin [(1-\lambda) \Omega]}{\sin \Omega} \cdot \delta^1+\frac{\sin [\lambda \Omega]}{\sin \Omega} \cdot \delta^2,
	\label{eq:slerp}
	\tag{\SLERP}
\end{equation}
where $\Omega$ is the angle between the two task vectors $\delta^1=\theta^1 - \theta_\mathrm{init}$ and $\delta^2=\theta^2 - \theta_\mathrm{init}$, and $\lambda$ the interpolation coefficient.
Critically \SLERP is applied layer by layer, each having a different angle. In \Cref{app:slerp:m} we clarify how \SLERP can be used iteratively to merge $M>2$ models.
To enforce diversity across weights, we simply vary the order in which text prompts $\vx$ are given in each run: this was empirically sufficient, though other diversity strategies could help, \eg varying the hyperparameters or the reward objectives (as explored in \Cref{app:fig:diversityLP_controlvskl}).

\begin{wrapfigure}[8]{hR!}{0.325\textwidth}
	\vspace{-4em}%
	\begin{center}
		\begin{tikzpicture}[x=1cm, y=1cm, thick, mid slashes/.style={decoration={markings,
						mark=at position 0.5 with {
								\draw[thick] (-3pt,-1pt) -- (0,2pt); %
								\draw[thick] (0,-1pt) -- (3pt,2pt); %
							}}, postaction={decorate}}, mid slash/.style={decoration={markings,
						mark=at position 0.5 with {
								\draw[-] (-2pt,-2pt) -- (2pt,2pt);
								\draw[-] (-2pt,2pt) -- (2pt,-2pt);
							}}, postaction={decorate}}
	]
	\node[ellipse, minimum height = 2cm, minimum width = 4.8cm, fill=colorgreenfull!10, , rotate=330] (v2) at (2,1.7) {};
	\node[ellipse, minimum height = 1cm, minimum width = 2cm, fill=colorgreenfull!20, rotate=315] (v1) at (1.5,2) {};

	\node[circle, fill=black, label=below:$\theta_\mathrm{init}$, inner sep=1.5pt] (init) at (0, 0) {};
	\node[circle, fill=black, label=above:$\theta^1$, inner sep=1.5pt] (rl1) at (0, 2) {};
	\node[circle, fill=black, label=below:$\theta^2$, inner sep=1.5pt] (rl2) at (4, 0) {};

	\pic [draw, angle eccentricity=-0.5, angle radius=0.5cm, "$\Omega\approx 90^\circ$"] {right angle = rl1--init--rl2};	

	\draw[->, very thick] (init.center) -- (rl1.center);
	\draw[->, very thick] (init.center) -- (rl2.center);

	\draw[-, very thick, dashed, red] (rl1.center) to[bend left=25] node[pos=0.4, above, black] {$\theta_\mathrm{slerp}^{\lambda}$} (rl2.center);

	\draw[-, very thick, dashed, colorbluefull] (rl1.center) -- (rl2.center) node[midway, right, black] {$\theta_\mathrm{lerp}^{\lambda}$};

	\path (rl1) to[bend left=25] coordinate[pos=0.377] (slerpMid) (rl2);
	\path (rl1) -- coordinate[pos=0.5] (lerpMid) (rl2);

	\draw[->, thick, dotted, colorredfull] (init.center) -- (slerpMid.center);
	\draw[->, thick, dotted, colorbluefull] (init.center) -- (lerpMid.center);

	\path[mid slashes] (rl1.center) -- (init.center) node[midway, left, black] {$\delta^{1}$};
	\path[mid slashes] (rl2.center) -- (init.center) node[midway, below, black] {$\delta^{2}$};

	\path[mid slash] (rl1.center) -- (lerpMid.center);
	\path[mid slash] (rl2.center) -- (lerpMid.center);
\end{tikzpicture}%
		\vspace{-1.8em}%
	\end{center}
	\caption{\SLERP \versus \LERP.}
	\label{fig:merge}
\end{wrapfigure}%

\textbf{Benefits from \SLERP \versus \LERP.}
Merging task vectors, either with \SLERP or \LERP, combines their abililities \cite{2022arXiv221204089I}.
The difference is that \SLERP preserves their norms, reaching higher rewards than the base models; this is summarized in \Cref{obs:slerp}.
In contrast, and as summarized in \Cref{obs:lerp}, the more standard \LERP has less impact on reward, but has the advantage of reducing \KL; indeed, as shown in \Cref{app:slerp:theoryslerpvslerp}, \LERP tends to pull the merged model towards the initialization, especially as the angle $\Omega$ between task vectors is near-orthogonal (see \Cref{obs:lerp}).
\begin{observation}[\SLERP]
	Spherical linear interpolation boosts rewards, yet slightly increases \KL. Empirical evidence in \Cref{sec:expe:slerp} and theoretical insights in \Cref{lemma:slerpkeepsnorm}.
	\label{obs:slerp}%
\end{observation}
\begin{observation}[\LERP]
	Linear interpolation reduces \KL, yet has reduced impact on reward. Empirical evidence in \Cref{app:slerp:slerpvslerplambda} and theoretical insights in \Cref{lemma:lerpreducesnorm,lemma:lerpreduceskl}.
	\label{obs:lerp}%
\end{observation}
\begin{observation}[Task vectors]
	Task vectors $\delta$ are close to orthogonal with $\Omega\approx90^\circ$, while the full weights $\theta$ are collinear. Empirical evidence in \Cref{app:slerp:angle}.
	\label{obs:tv}%
\end{observation}
\subsection{\textcolor{colorredfull}{Stage 3: linear interpolation towards initialization}}
\label{sec:model:liti}
\textbf{\LITI.}
In the previous stage, \SLERP combines multiple policies into one with higher rewards and slightly higher \KL.
This third stage, inspired by \WISE from \citet{Wortsman2022robust}, interpolates from the merged model towards the initialization:%
\begin{equation}%
	\theta^{\eta} \leftarrow (1-\eta) \cdot \theta_\mathrm{init} + \eta \cdot \theta_\mathrm{slerp}.%
	\label{eq:liti}%
	\tag{\LITI}%
\end{equation}%
Adjusting the interpolating coefficient $\eta\in[0,1]$ trades off between some newly acquired behaviors leading to high rewards \versus general knowledge from the \SFT initialization.
Specifically, large values $\eta \approx 1$ provide high rewards but also high \KL, while smaller values $\eta \approx 0$ lean towards smaller rewards and minimal \KL.
Fortunately, we observe that the reduction in \KL is proportionally greater than the reduction in reward when decreasing $\eta$.
Then, \LITI empirically yields Pareto fronts that are noticeably above the \enquote{diagonal}, but also above those revealed during the base RLs.
\begin{observation}[\LITI]%
	Interpolating weights towards the initialization reveals a better Pareto front than the one revealed during RL fine-tuning. Empirical evidence in \Cref{fig:main:paretofront} and \Cref{sec:expe:liti}, and theoretical insights in \Cref{lemma:klupperbound,lemma:liti-pareto}.%
	\label{obs:liti}%
\end{observation}%
\textbf{Benefits from \LITI.}
Previous works tried to understand how weight interpolation can mitigate forgetting while increasing robustness and generalization.
\cite{Wortsman2022robust} argues that \WISE, akin to \LITI in supervised learning contexts, recovers generalizable features from pre-training that might be lost during fine-tuning \cite{kumar2022finetuning}, consistently with \WA reducing catastrophic forgetting \cite{stojanovski2022momentum,eeckt2022weight} in continual learning.
Then in the context of RL, \cite{lin2024mitigating} argues that \LITI increases feature diversity, efficiently balancing between generality and task specificity.
Finally, \cite{jang2024model} argues that the geometric projection of the ideal weights is located between the merged model and the initialization.
\subsection{\textcolor{colorgreenfull}{Iterative \WARP}}
\label{sec:model:warpi}
\textbf{Iterative training.}
The model merging strategies previously described not only establish an improved Pareto front of solutions, but also set the stage for iterative improvements. Indeed, if the computational budget is sufficient, we can apply those three stages iteratively, using $\theta^{\eta}$ from previous Pareto front (usually with $\eta=0.3$, choice ablated in \Cref{app:expe:ablationeta}) as the initialization $\theta_\mathrm{init}$ for the next iteration, following the model recycling \cite{choshen2022cold,rame2022recycling} strategies.
Then, the entire training procedure is made of multiple iterations, each consisting of those three stages, where the final weight from a given iteration serves as an improved initialization for the next one.
\begin{observation}[Iterative \WARP]%
	Applying \WARP iteratively improves results, converging to an optimal Pareto front. Empirical evidence in \Cref{sec:expe:warpi,sec:expe:sxs}.%
	\label{obs:warpi}%
\end{observation}%

\section{Experiments: on the benefits of \WARP}
\label{sec:expe}

\textbf{Setup.}
We consider the \GEMMAB \cite{team2024gemma} LLM, which we seek to fine-tune with \RLHF into a better conversational agent.
We use REINFORCE~\cite{williams1992simple} policy gradient to optimize the \KL-regularized reward.
The dataset $\X$ contains conversation prompts.
We generate on-policy samples with temperature $0.9$, batch size of $128$, Adam~\cite{kingma2014adam} optimizer with learning rate $10^{-6}$ and warmup of $100$ steps.
\SLERP is applied independently to the 28 layers.
Except when stated otherwise, we train for $T=9k$ steps, with \KL strength $\beta=0.1$, \EMA update rate $\mu=0.01$, merging $M=2$ policies uniformly $\lambda=0.5$, and \LITI update rate $\eta=0.3$; we analyze those values in \Cref{app:expe}.
We rely on a high capacity reward model, the largest available, which prevents the use of an oracle control RM as done in \cite{gao2022scaling,rame2024warm}.

\textbf{Summary.}
In our experiments, we analyze the \KL to the \SFT policy (reflecting the forgetting of pre-trained knowledge) and the reward (evaluating alignment to the RM).
In \Cref{sec:expe:ema}, we first show the benefits of using an \EMA anchor; then in \Cref{sec:expe:slerp}, we show that merging policies trained independently helps.
Moreover, in \Cref{sec:expe:liti}, we show that \LITI improves the \KL-reward Pareto front; critically, repeating those three \WARP stages can iteratively improve performances in \Cref{sec:expe:warpi}.
A limitation is that our RM accurately approximates true human preferences only in low \KL region, though can be hacked away from the SFT \cite{gao2022scaling}.
Therefore, we finally report other metrics in \Cref{sec:expe:sxs}, specifically comparing against open-source baselines such as Mixtral \cite{jiang2024mixtral}, and reporting performances on standard benchmarks such as MMLU~\citep{hendrycks19measuring}.
\subsection{\textcolor{colorbluefull}{Stage 1: exponential moving average as a dynamic anchor in \KL regularization}}%
\label{sec:expe:ema}%
\begin{figure*}[b!]
	\begin{center}
		\begin{subfigure}[b]{0.325\textwidth}
			\includegraphics[width=\textwidth]{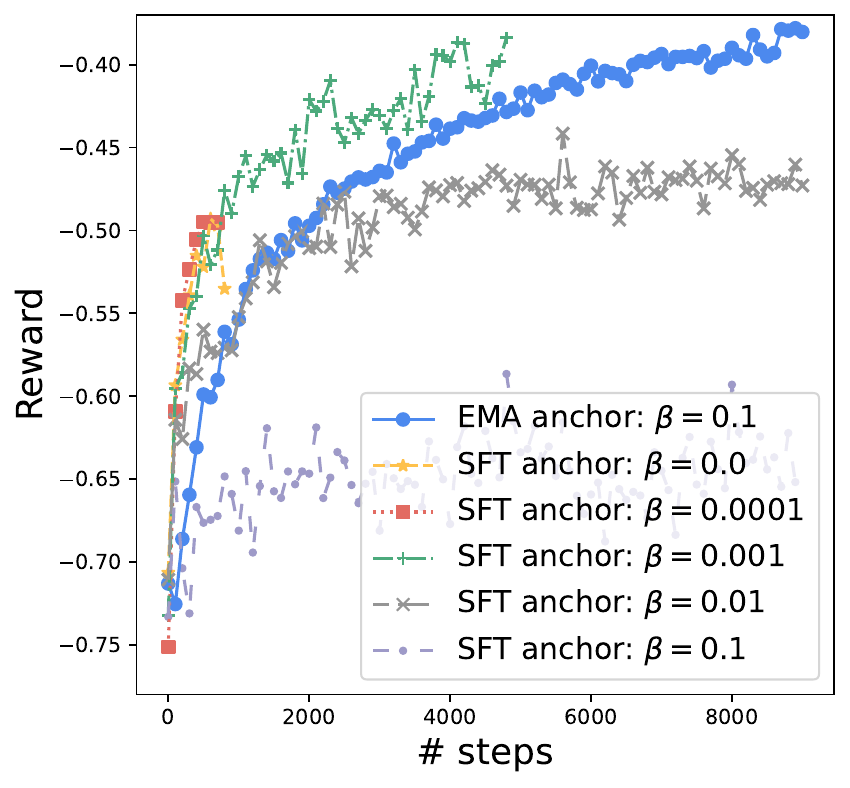}
			\caption{Reward \versus steps.}
			\label{fig:contremalp_controlvsstep}%
		\end{subfigure}%
		\hfill
		\begin{subfigure}[b]{0.325\textwidth}%
			\includegraphics[width=\textwidth]{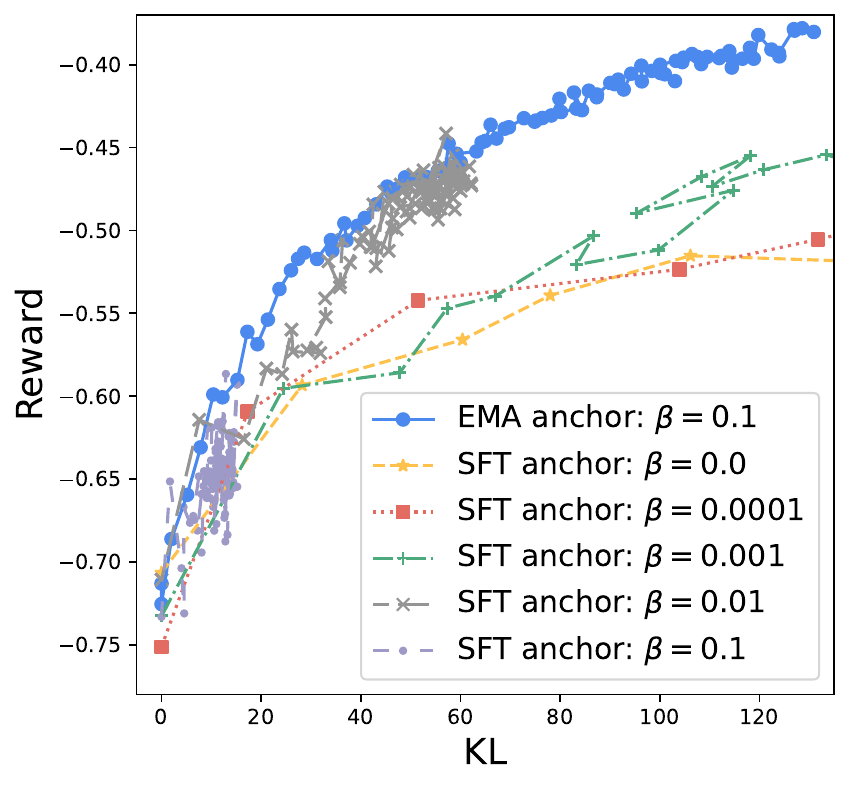}
			\caption{Reward \versus \KL.}%
			\label{fig:anchor_ema_vs_sft}%
		\end{subfigure}
		\hfill
		\begin{subfigure}[b]{0.325\textwidth}
			\includegraphics[width=1.0\textwidth]{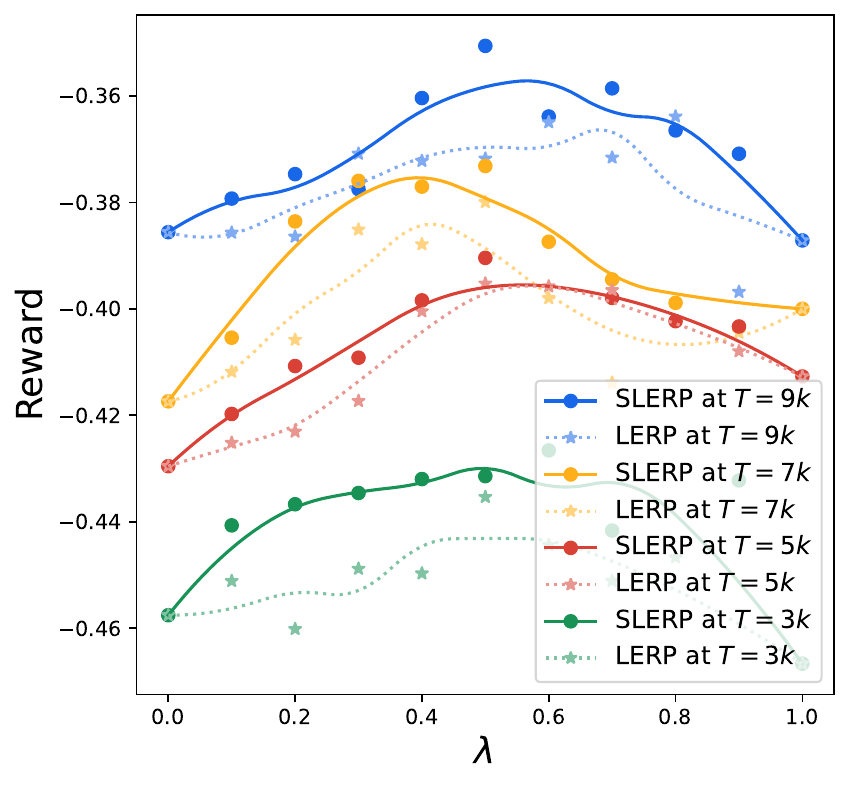}
			\caption{\SLERP \versus \LERP.}
			\label{fig:interpolation8090_controlvslambda}%
		\end{subfigure}%
		\hfill
	\end{center}%
	\caption{\textbf{\EMA and \SLERP experiments.} We first compare RL runs with different anchors and strengths $\beta$ in the \KL regularization. We show their results along training in \Cref{fig:contremalp_controlvsstep}, and their \KL-reward Pareto fronts in \Cref{fig:anchor_ema_vs_sft}. We perform evaluation every $100$ steps, and train them for $T=9k$ steps, though we stopped the trainings if they ever reach a \KL of $200$ (\eg after $T=1k$ training steps when $\beta=0.0$).
		\Cref{fig:interpolation8090_controlvslambda} plots the reward obtained when merging two policies (trained independently after $T$ RL steps with their own \EMA anchor) with interpolating coefficient $\lambda$; highest rewards are with \SLERP for $\lambda=0.5$ and $T=9k$ steps.}
	\label{fig:expes_ema}%
\end{figure*}%
In \Cref{fig:contremalp_controlvsstep,fig:anchor_ema_vs_sft}, we compare the training trajectories of different REINFORCE variants, where the changes lie in the choice of the anchor in the \KL regularization and of the hyperparameter $\beta$ controlling its strength.
Results are computed every 100 training steps.
In our proposed version, the anchor is the \EMA of the trained policy with $\beta=0.1$ and an \EMA update rate $\mu=0.1$ (other values are ablated in \Cref{app:fig:expes_ablation}).
As the Pareto front for our strategy is above and to the left in \Cref{fig:anchor_ema_vs_sft}, this confirms the superiority of using such an adaptive anchor.
The baseline variants all use the \SFT as the anchor, with different values of $\beta$.
The lack of regularization ($\beta=0.0$) leads to very fast optimization of the reward in \Cref{fig:contremalp_controlvsstep}, but largely through hacking, as visible by the \KL exploding in just a few training steps in \Cref{fig:anchor_ema_vs_sft}.
In contrast, higher values such as $\beta=0.1$ fail to optimize the reward as regularization is too strong, causing a quick reward saturation around $-0.62$ in \Cref{fig:contremalp_controlvsstep}. Higher values such as $\beta=0.01$ can match our \EMA anchor in low \KL regime, but saturates around a reward of $-0.46$.
In contrast, as argued in \Cref{obs:ema}, the dynamic \EMA anchor progressively moves away from the SFT initialization, causing implicit annealing of the regularization.
In conclusion, relaxing the anchor with \EMA updates allows the efficient learning of \KL-reward Pareto-optimal policies, at any given \KL level, for a fixed compute budget.
We refer the interested reader to additional experiments in \Cref{fig:emateacher_vs_student} from \Cref{app:expe:ablationmubeta} where we compare the trained policies with their online \EMA version.
\FloatBarrier
\subsection{\textcolor{coloryellowfull}{Stage 2: spherical linear interpolation of independently rewarded policies}}
\label{sec:expe:slerp}
In \Cref{fig:interpolation8090_controlvslambda}, we plot $\lambda \rightarrow r\left( \operatorname{slerp}\left(\theta_\mathrm{init}, \theta^1, \theta^2, \lambda\right)\right)$ showing reward convexity when interpolating policies via \SLERP, validating \Cref{obs:slerp}.
This mirrors the linear mode connectivity~\cite{Frankle2020} property across weights fine-tuned from a shared initialization, \ie the fact that interpolated weights perform better than the initial models (recovered for $\lambda=0$ or $\lambda=1$).
Moreover, \SLERP consistently obtains higher rewards than \LERP;
yet, this is at slightly higher \KL, as further detailed in 
\Cref{app:slerp:theoryslerpvslerp,app:slerp:slerpvslerplambda}, where we analyze respectively their theoretical and empirical differences.
\FloatBarrier
\subsection{\textcolor{colorredfull}{Stage 3: linear interpolation towards initialization}}
\label{sec:expe:liti}
\begin{figure*}[b!]
	\begin{center}
		\begin{subfigure}[b]{0.325\textwidth}
			\includegraphics[width=1.0\textwidth]{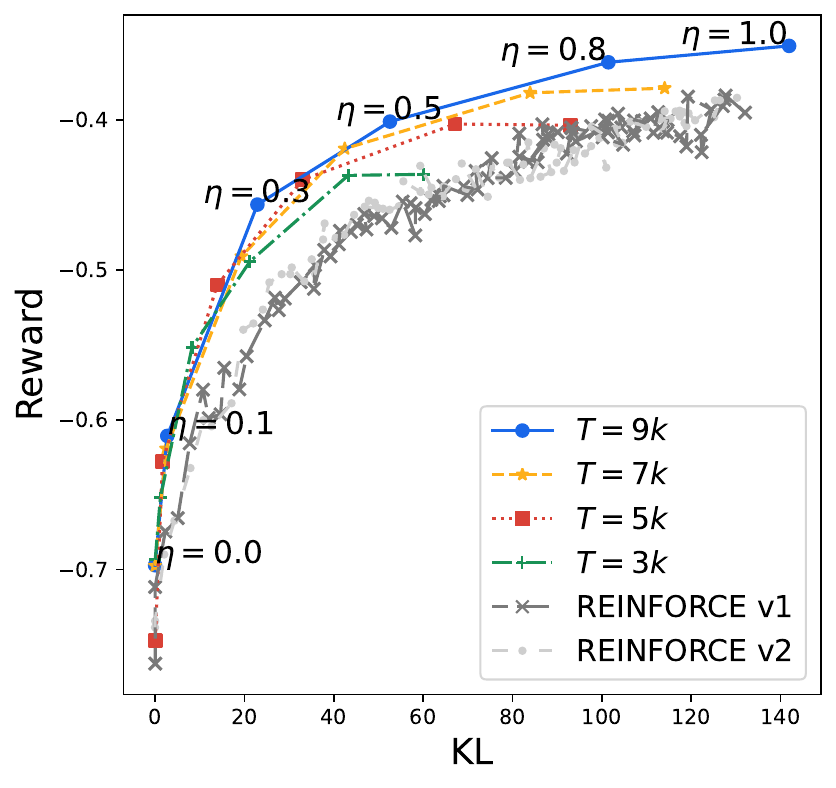}
			\caption{\LITI of \SLERP after $T$ steps.}
			\label{fig:num_steps}%
		\end{subfigure}%
		\hfill
		\begin{subfigure}[b]{0.325\textwidth}%
			\includegraphics[width=\textwidth]{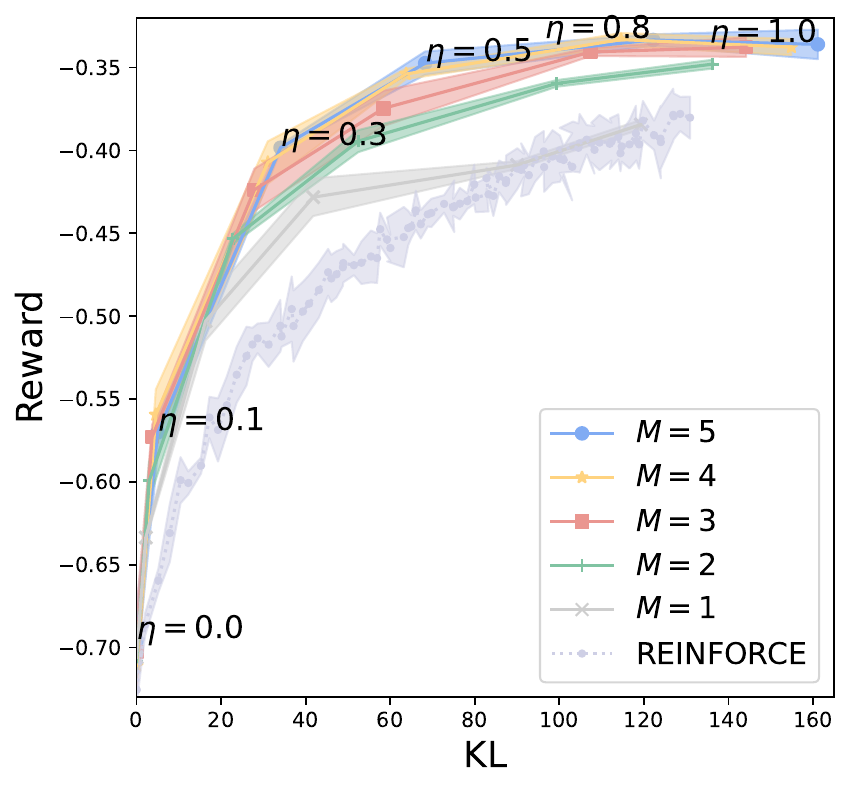}
			\caption{\LITI of \SLERP of $M$ weights.}%
			\label{fig:impact_m}%
		\end{subfigure}
		\hfill
		\begin{subfigure}[b]{0.325\textwidth}%
			\includegraphics[width=\textwidth]{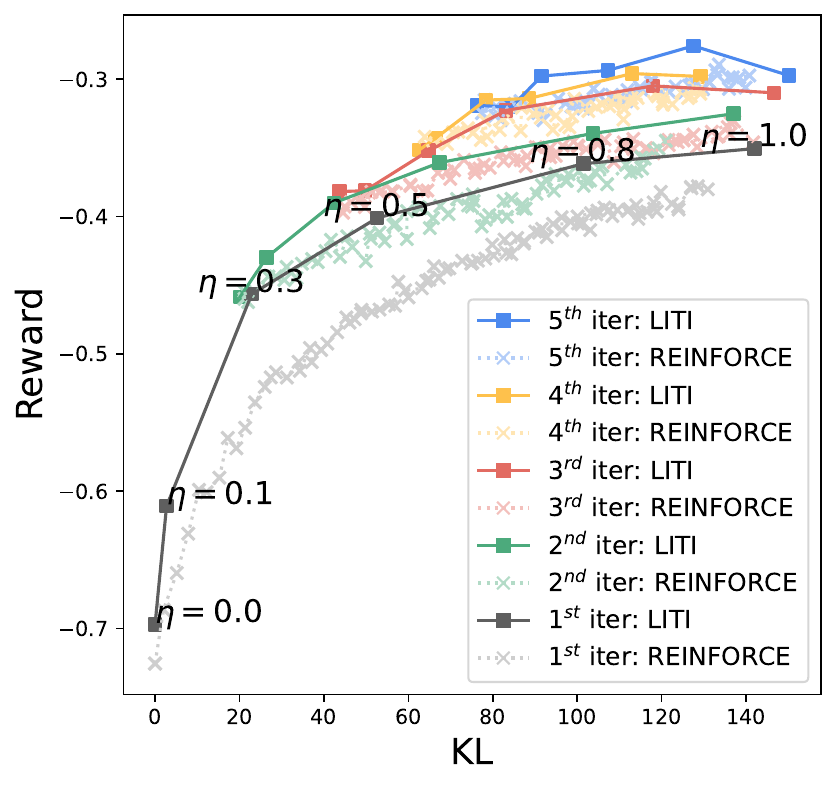}
			\caption{Iterative \WARP.}%
			\label{fig:warp_vs_iterations}
		\end{subfigure}%
	\end{center}%
	\caption{\textbf{\LITI and iterative experiments.}
		\Cref{fig:num_steps} considers the \LITI of the \SLERP of $M=2$ policies after $T$ steps with $\lambda=0.5$, interpolating towards their \SFT init as we slide $\eta$, revealing Pareto fronts above the $M=2$ REINFORCE training trajectories.
		Then \Cref{fig:impact_m} plots the \LITI of the \SLERP of $M$ weights with $\lambda=\frac{1}{M}$ after $T=9k$ steps: light-colored areas show standard deviations across $5$ experiments.
		The iterative \WARP procedure is illustrated in \Cref{fig:warp_vs_iterations}; we fine-tune $M=2$ policies with their own \EMA as the anchor, merge them with \SLERP, interpolate towards their init with \LITI, and iteratively leverage the weights obtained with $\eta=0.3$ as the new initialization for the next iteration.}%
	\label{fig:expes_slerp_liti}%
\end{figure*}

In \Cref{fig:num_steps}, we merge policies trained for $T$ steps, and then apply the \LITI procedure. Critically, sliding the interpolating coefficient $\eta\in\{0,0.1, 0.3, 0.5, 0.8, 1.0\}$ reveals various Pareto fronts, consistently above the training trajectories obtained during the two independent RL fine-tunings.
Interestingly, longer fine-tunings improve performances, at high \KL, but also at lower \KL, simply by using a smaller $\eta$ afterwards.
Then in \Cref{fig:impact_m}, we report the Pareto fronts when merging up to $M=5$ weights.
We note that all Pareto fronts revealed when applying \LITI are consistently above the ones from RL fine-tunings, validating \Cref{obs:liti}.
More precisely, best results are achieved by merging an higher number of policies $M$, suggesting a promising scaling direction.

\subsection{\textcolor{colorgreenfull}{Iterative \WARP}}
\label{sec:expe:warpi}
In \Cref{fig:warp_vs_iterations}, we apply the iterative procedure described in \Cref{sec:model:warpi}.
At each of the $I=5$ iterations we train $M=2$ policies for $T$ steps, with $T=9k$ for the first iteration, and $T=7k$ for iterations 2 and 3, and then $T=5k$ for computational reasons.
The \LITI curves interpolate towards their own initialization (while \Cref{fig:main:paretofront} interpolated towards the SFT initialization, see \Cref{app:expe:ablationinit} for a comparison).
We systematically observe that \LITI curves are above the RL training trajectories used to obtain the inits.
Results get better at every iteration, validating \Cref{obs:warpi}, although with reduced returns after a few iterations.
\FloatBarrier
\clearpage
\subsection{Comparisons and benchmarks}
\label{sec:expe:sxs}
\textbf{Side by side comparisons.}
To conclude our experiments, we compare our trained policies against Mistral \cite{jiang2023mistral} and Mixtral \cite{jiang2024mixtral} LLMs.
Each policy generates a candidate answer on an held-out collection of prompts, as in the Gemma tech report \cite{team2024gemma}.
Then similarly to Gemini 1.5 \cite{reid2024gemini}, we compute side by side preference rates \cite{zheng2023judging} with \enquote{much better}, \enquote{better} and \enquote{slightly better} receiving scores of $\pm1.5$, $\pm1$, and $\pm0.5$ respectively (and ties receiving a score of $0$).
A positive score represents better policies.
The results in \Cref{tab:sxs} validate the efficiency of \WARP, as our policies are preferred over Mistral variants, and also outperform the two previous \GEMMAB releases. However, we note that the results stagnate after the \nth{3} iteration.%
\begin{table}[t!]%
	\centering%
	\caption{\textbf{Side by side comparisons}.}%
	\centering
	\adjustbox{max width=0.70\textwidth}{%
		\begin{tabular}{lccc}%
			\toprule
			\textbf{Methods}   & \textbf{Mistral 7B v1} & \textbf{Mistral 7B v2} & \textbf{Mixtral 8x7B} \\
			\midrule
			\GEMMAB 1.0           & 0.24                     & -0.01                    & -0.08 \\
			\GEMMAB 1.1           & 0.37                     & 0.16                     & 0.08 \\
			REINFORCE \EMA anchor & 0.37                     & 0.16                     & 0.07 \\
			\WARP: \nth{1} iter   & 0.42                     & 0.23                     & 0.13 \\
			\WARP: \nth{2} iter   & \textbf{0.45}                     & 0.25                     & 0.16 \\
			\textbf{\WARP: \nth{3} iter}   & \textbf{0.45}                     & \textbf{0.26}                     & \textbf{0.18}      \\   
			\WARP: \nth{4} iter   & \textbf{0.45}                     & 0.25                     & 0.16 \\
			\WARP: \nth{5} iter   & \textbf{0.45}                     & 0.24                     & 0.17 \\
			\bottomrule%
		\end{tabular}}%
	\label{tab:sxs}%
\end{table}%

\textbf{Benchmarks.}
\Cref{tab:bench} compares \WARP (\nth{3} iter) and the latest \GEMMAB 1.1 release \cite{team2024gemma} on popular benchmarks in the zero-shot setup: MBPP~\citep{austin21program} and HumanEval~\citep{chen21evaluating} benchmarking coding capabilities, MMLU~\citep{hendrycks19measuring} assessing STEM knowledge, the GSM8K~\citep{cobbe21training} and MATH~\citep{hendrycks2021measuring} benchmarks targeting reasoning abilities, and the Big Bench Hard (BBH)~\citep{suzgun2022challenging} benchmark evaluating general capabilities through questions that were deemed difficult for frontier LLMs.
\WARP has particularly strong results on mathematics benchmarks, suggesting higher analytical capabilities.%
\begin{table}[h!]%
	\centering%
	\caption{\textbf{Benchmark results}.}%
	\centering%
	\adjustbox{max width=0.7\textwidth}{%
		\begin{tabular}{lcccccc}%
			\toprule%
			\textbf{Methods} & MBPP & MMLU & GSM8K & MATH & HumanEval & BBH  \\%
			\midrule%
			\GEMMAB 1.1      & 39.0 & 56.4 & 55.6  & 25.6 & 46.9      & 53.1 \\%
			\textbf{\WARP}            & \textbf{45.4} & \textbf{57.6} & \textbf{66.8}  & \textbf{31.0} & \textbf{50.0}      & \textbf{58.8} \\%
			\bottomrule%
		\end{tabular}%
	}%
	\label{tab:bench}%
\end{table}%

\FloatBarrier%

\section{Related work}%
\label{sec:related}
\textbf{How to merge models.}
The question of how best to merge models has recently garnered significant attention, driven by the discoveries that deep models can be merged in the weight space \cite{utans1996weight} instead of in the prediction space, as traditionally done in ensembling \cite{hansen1990neural,Lakshminarayanan2017}. For clarity, we collectively refer to these methods as weight averaging (\WA).
The most common is \LERP, initially used to average checkpoints collected along a single run, uniformly \cite{szegedy2016rethinking,izmailov2018} or with an exponential moving average (\EMA)~\cite{polyak1992acceleration}, notably as a mean teacher~\cite{tarvainen2017mean} for self-supervision \cite{sohn20_fixmat,he2020momentum,oquab2024dinov,Caron_2021_ICCV,grill2020bootstrap}.
Following the linear mode connectivity~\cite{Frankle2020} observation, the model soups variants \cite{Wortsman2022ModelSA,2022arXiv221204089I,rame2022recycling} linearly interpolate from different fine-tunings; this relies on the shared pre-training, limiting divergence \cite{Neyshabur2020} such as models remain in constrained weight regions \cite{gueta2023knowledge}, which also suggests that pre-training mitigates the need to explicitly enforce trust regions in gradient updates~\cite{schulman2015trust, schulman2017proximal}.
Subsequent works such as TIES merging \cite{yadav2023tiesmerging} and DARE \cite{yu2023language} reduce interferences in multi-task setups with sparse task vectors~\cite{2022arXiv221204089I}. In contrast, we use \SLERP, introduced in \cite{shoemake1985animating}, increasingly popular in the open-source community \cite{goddard2024arcee} but relatively underexplored in the academic literature, with limited studies such as \cite{kim2024token}.
Some tried to align weights trained from scratch \cite{entezari2022the, ainsworth2022gitrebasin} or with different architectures \cite{wan2024knowledge}; yet, the methods are complex, less robust, and usually require additional training.

\textbf{Benefits of model merging.} \WA boosts generalization by reducing variance \cite{Wortsman2022ModelSA, rame2022diwa}, decreasing memorization \cite{lin2023spurious, zaman2023fuse, rame2024warm} and flattening the loss landscape \cite{cha2021wad}.
Additionally, merging weights combines their strengths \cite{2022arXiv221204089I}, which helps in multi-task setups \cite{ilharco2022patching, rame2023rewarded}, to tackle catastrophic forgetting \cite{stojanovski2022momentum, eeckt2022weight} or to provide better initializations \cite{choshen2022cold}, as explored in \cite{jain2023dart,jang2024model,huang2024imwa} for iterative procedures in classification tasks.
In particular, we considered using the geometric insights from Eq. 2 in \cite{jang2024model}; yet, as our task vectors are nearly orthogonal $\Omega \approx 90^\circ$ (see \Cref{app:slerp:angle}), using the update rule $ \eta \rightarrow \frac{2 \cos \Omega}{1+\cos \Omega} $ failed.
\WA is now also used in RL setups \cite{nikishin2018improving,gaya2021learning,lawson2023merging}; for example, \WARM \cite{rame2024warm} merges reward models to boost their efficiency, robustness and reliability.
Actually, \WARP is conceived as a response to \WARM, demonstrating that model merging can tackle two key \RLHF challenges; policy learning in \WARP and reward design in \WARM.
The most similar works are the following, which also explore how \WA can improve policy learning.
\cite{noukhovitch2023language} proposes an iterative approach with the \EMA as a new initialization for subsequent iterations.
\cite{gorbatovski2024learn} and \cite{munos2023nash} uses \EMA as the reference, but only for direct preference optimization.
\cite{rame2023rewarded} employs \LERP to improve alignment in multi-objective RLHF when dealing with different objectives; similarly, \cite{xiao2023lm} targets multi-task setups with \LERP.
Finally, \cite{lin2024mitigating} and \cite{fu2024disperse} use model merging to reduce the alignment tax, although without incorporating \EMA during training, without merging multiple rewarded policies and not iteratively.
Critically, none of these works focus on \KL as a measure of forgetting, use \EMA as the anchor in \KL, apply \SLERP or use \LITI as the initialization for subsequent RL iterations.
In contrast, \WARP integrates all those elements, collectively leading to an LLM outperforming Mixtral \cite{jiang2024mixtral}.

\section{Discussion}%
\label{sec:discussion}%
\textbf{Distributed learning for parallelization and open-source.}
\WARP addresses a crucial challenge: aligning LLMs with human values and societal norms, while preserving the capabilities that emerged from pre-training. To this end, we leverage a (perhaps surprising) ability: policies trained in parallel can combine their strengths within a single policy by weight averaging.
Then, the distributed nature of \WARP makes it flexible and scalable, as it is easily parallelizable by enabling intermittent weight sharing across workers.
Actually, iterative \WARP shares similarities with DiLoCo \cite{douillard2023diloco}: by analogy, the first stage performs inner optimization on multiple workers independently; the second stage merges gradients from different workers; the third stage performs SGD outer optimization with a learning rate equal to $\eta$.
More generally, \WARP could facilitate open-source \cite{goddard2024arcee} collaborative training of policies \cite{updatablemachinelearning}, optimizing resource and supporting privacy in federated learning \cite{mcmahan2017communication} scenarios; collaborators could train and share their LLMs, while keeping their data and RMs private.
In particular, we show in \Cref{app:length_penalty} that \WARP can handle diverse objectives, similarly to \cite{rame2023rewarded}.

\textbf{Iterated amplification.}
\WARP improves LLM alignment by leveraging the principles of iterated amplification \cite{christiano2018supervising} and progressive collaboration of multiple agents. By analogy, model merging via \WA acts as an effective alternative to debate \cite{irving2018ai}, with agents communicating within the weight space instead of the token space, ensuring that only essential information is retained \cite{rame2024warm}.
Then, \WARP refines the training signal by combining insights and exploration from diverse models, iteratively achieving higher rewards through self-distillation \cite{tarvainen2017mean}, surpassing the capabilities of any single agent. If this is the way forward, then an iterative safety assessment would be required to detect and mitigate potential risks early, ensuring that the development remains aligned with safety standards.
\clearpage

\textbf{Scaling alignment.}
The \WARP procedure increases the compute training cost by performing multiple fine-tunings at each iteration.
Yet, this should be viewed as \enquote{a feature rather than a bug}. Specifically, by preventing memorization and forgetting, we see \WARP as a fine-tuning method that can transform additional compute allocated to alignment into enhanced capabilities and safety. This would allow scaling (the traditionally cheap) post-training alignment, in the same way pre-training has been scaled \cite{hoffmann2022training}.
Critically for large-scale deployment, the acquired knowledge is within a single (merged) model, thus without inference or memory overhead, in contrast to \enquote{more agents} approaches \cite{li2024more,wang2024mixture}.
Finally, although \WARP improves policy optimization, it is important to recognize that \WARP does not address other critical challenges in \RLHF \cite{casper2023open}: to mitigate the safety risks~\cite{amodei2016concrete,hendrycks2022x,hendrycks2023natural} from misalignment~\cite{taylor2016alignment,ngo2022alignment}, \WARP should be part of a broader responsible AI framework.%

\section{Conclusion}%
\label{sec:conclusion}%
We introduce Weight Averaged Rewarded Policies (\WARP), a novel \RLHF strategy to align LLMs with three distinct stages of model merging: exponential moving average as a dynamic anchor during RL, spherical interpolation to combine multiple policies rewarded independantly, and interpolation towards the shared initialization.
This iterative application of \WARP improves the \KL-reward Pareto front, aligning the LLMs while protecting the knowledge from pre-training, and compares favorably against state-of-the-art baselines.
We hope \WARP could contribute to safe and powerful AI systems by scaling alignment, and spur further exploration of the magic behind model merging.

\clearpage
\newpage
\bibliographystyle{plainnat}
\bibliography{main}

\clearpage
\newpage
\appendix
\hrule
\begin{center}
    \Large \WARP: On the Benefits of Weight Averaged Rewarded Policies
\end{center}

\begin{center}
    \large Supplementary material
\end{center}
\hrule
\vskip 0.5cm
This supplementary material is organized as follows:
\begin{itemize}
    \item \Cref{app:details} provides additional illustration of the \WARP procedure.
    \item \Cref{app:slerp:theoryslerpvslerp} details theoretical insights on task vectors, \SLERP, \LERP and \LITI.    
    \item \Cref{app:slerp:empirical} details empirical insights on task vectors, \SLERP, \LERP and \LITI.
    \item \Cref{app:expe} shows the impact of different design choices in \WARP.
    \item \Cref{app:length_penalty} investigates a potential length bias in \WARP, and how to fix it.
    \item \Cref{app:dlp} explores the relationship between \KL and diversity in generations. 
\end{itemize}
\FloatBarrier
\section{Strategy illustration}
\label{app:details}
\begin{figure*}[b!]
	\begin{center}
		\resizebox{0.9\textwidth}{!}{\begin{tikzpicture}[x=1cm, y=1cm]
	\newcommand{\YL}{8.5}
	\newcommand{\YG}{9}
	\draw[->, very thick, black]           (\YL, -1) -- (\YG, -1) node [pos=1, right, sloped]     (lab1) {REINFORCE~\cite{williams1992simple}};
	\draw[<->, very thick, dotted, black]           (\YL, -1.5) -- (\YG, -1.5) node [pos=1, right, sloped]     (lab1) {\KL~\cite{jaques2017sequence}};
	\draw[->, very thick, dashed, colorbluefull]           (\YL, -2) -- (\YG, -2) node [pos=1, right, sloped] (lab1) {\EMA \cite{izmailov2018}};
	\draw[->, very thick, dashed, coloryellowfull]           (\YL, -2.5) -- (\YG, -2.5) node [pos=1, right, sloped] (lab1) {\SLERP \cite{shoemake1985animating} of task vectors \cite{2022arXiv221204089I}};
	\draw[->, very thick, dashed, colorredfull]           (\YL, -3) -- (\YG, -3) node [pos=1, right, sloped] (lab1) {\LITI \cite{Wortsman2022robust}};

	\newcommand{\YY}{3}
	\newcommand{\YZ}{2}
	\newcommand{\YA}{0}
	\newcommand{\YB}{-2}
	\newcommand{\YC}{-3}
	\node[circle, fill=colorredfull!20] (init1)    at (0, \YA) {$\theta_\mathrm{init}$};
	\node[circle, fill=colorredfull!20] (init2)    at (3, \YA) {$\theta_\mathrm{init}^{\prime}$};

	\node[circle, draw] at (1, 0.5) (rl111) {};
	\node[circle, draw] at (2, 1.0) (rl112) {};
	\node[circle, draw] at (3, 1.5) (rl113) {};

	\node[circle, fill=colorbluefull!20] (ema11) at (2, \YY) {$\theta_\mathrm{ema}^1$};

	\node[circle, draw] at (1, -0.5) (rl121) {};
	\node[circle, draw] at (2, -1.0) (rl122) {};
	\node[circle, draw] at (3, -1.5) (rl123) {};

	\node[circle, fill=colorbluefull!20] (ema12) at (2, \YC) {$\theta_\mathrm{ema}^2$};

	\node[circle, fill=black!20] (rl11) at (4, \YZ) {$\theta_\mathrm{rl}^1$};
	\node[circle, fill=black!20] (rl12) at (4, \YB) {$\theta_\mathrm{rl}^2$};
	\node[circle, fill=coloryellowfull!20] (slerp1) at (5, \YA) {$\theta_\mathrm{slerp}$};
	\node[circle, fill=coloryellowfull!20] (slerp2) at (8, 0) {$\theta_\mathrm{slerp}^{\prime}$};

	\node[circle, fill=black!20] (rl21) at (7, \YZ) {...};
	\node[circle, fill=black!20] (rl22) at (7, \YB) {...};

	\draw[->, very thick, black] (init1) -- (rl11);%
	\draw[->, very thick, black] (init1) -- (rl12);%

	\draw[<->, very thick, dotted, black] (ema11) -- (rl11);
	\draw[<->, very thick, dotted, black] (ema12) -- (rl12);

	\draw[->, very thick, dashed, colorbluefull] (init1) -- (ema11);
	\draw[->, very thick, dashed, colorbluefull] (rl111) -- (ema11);
	\draw[->, very thick, dashed, colorbluefull] (rl112) -- (ema11);
	\draw[->, very thick, dashed, colorbluefull] (rl113) -- (ema11);

	\draw[->, very thick, dashed, colorbluefull] (init1) -- (ema12);
	\draw[->, very thick, dashed, colorbluefull] (rl121) -- (ema12);
	\draw[->, very thick, dashed, colorbluefull] (rl122) -- (ema12);
	\draw[->, very thick, dashed, colorbluefull] (rl123) -- (ema12);

	\draw[->, very thick, dashed, coloryellowfull] (rl11) to [bend left=20] (slerp1);
	\draw[->, very thick, dashed, coloryellowfull] (rl12) to [bend right=20] (slerp1);

	\draw[->, very thick, dashed, colorredfull] (slerp1) -- (init2);
	\draw[->, very thick, dashed, colorredfull] (init1) to (init2);

	\draw[->, very thick, black] (init2) -- (rl21);
	\draw[->, very thick, black] (init2) -- (rl22);

	\draw[->, very thick, dashed, coloryellowfull] (rl21) to [bend left=20] (slerp2);
	\draw[->, very thick, dashed, coloryellowfull] (rl22) to [bend right=20] (slerp2);

	\draw[->, very thick, dashed, colorredfull] (slerp2) edge[bend right=25] node[midway, below, color=colorredfull]{$(1-\eta) \cdot \theta_\mathrm{init} + \eta \cdot \theta_\mathrm{slerp}^{\prime}$} (init1);

\end{tikzpicture}}%
	\end{center}%
	\caption{
		\textbf{Detailed illustration of the \WARP strategy}.
		From a (pre-trained and supervised fine-tuned) LLM $\theta_\mathrm{init}$, we launch $M=2$ fine-tunings (black arrows \protect\tikz[baseline]\protect\draw[->, very thick, black](0ex,0.8ex) -- (3ex,0.8ex);).
		The innovation of \WARP lies in the use of model merging by weight averaging at three different stages.
		First, the exponential moving averages (\EMA, blue dashed arrows \protect\tikz[baseline]\protect\draw[->, very thick, dashed, colorbluefull](0ex,0.8ex) -- (3ex,0.8ex);) of the policy (collected at different training steps) serves as the anchor for the \KL regularization (black double-headed dotted arrows \protect\tikz[baseline]\protect\draw[<->, very thick, dotted, black](0ex,0.8ex) -- (3ex,0.8ex);).
		The fine-tuned networks are weight averaged using spherical linear interpolation of task vectors (\SLERP, yellow dashed arrows \protect\tikz[baseline]\protect\draw[->, very thick, dashed, coloryellowfull](0ex,0.8ex) -- (3ex,0.8ex);).
		Third, we interpolate towards the initialization (\LITI, red dashed arrows \protect\tikz[baseline]\protect\draw[->, very thick, dashed, colorredfull](0ex,0.8ex) -- (3ex,0.8ex);).
		This obtained model $\theta_\mathrm{init}^{\prime}$ serves as an updated initialization for the next iteration, progressively refining the model’s capabilities and alignment.
		Overall, the final model $\theta_\mathrm{slerp}^{\prime}$ has high reward but also high \KL.
		Then, by interpolation towards the \SFT init, we reveal a \KL-reward Pareto front of solutions: $\{(1-\eta) \cdot \theta_\mathrm{sft} + \eta \cdot \theta_\mathrm{slerp}^{I}\mid0\leq\eta\leq1\}$.
	}%
	\label{fig:warp_tikz}%
\end{figure*}
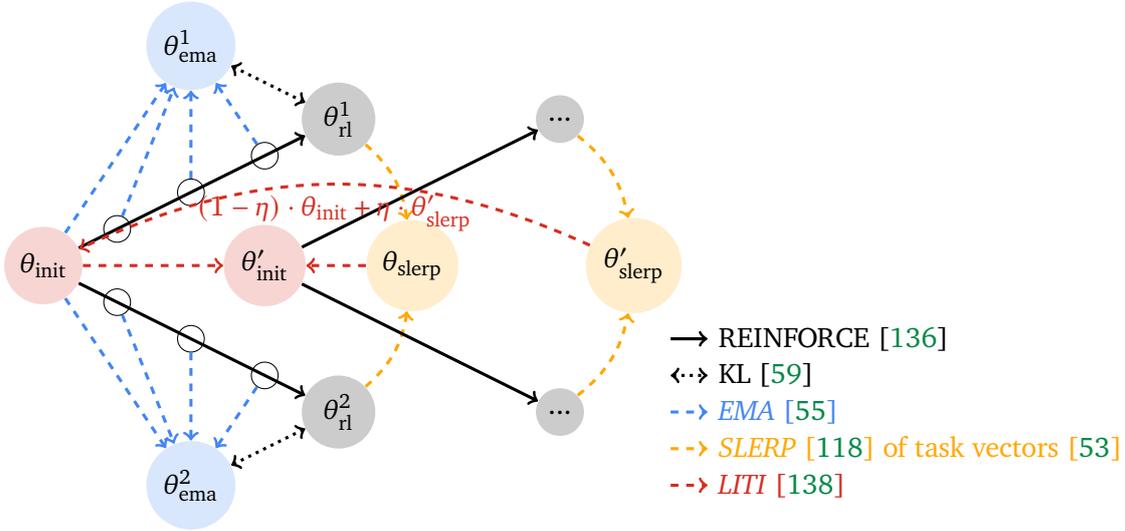

\begin{figure*}[t!]
	\begin{center}
		\resizebox{0.8\textwidth}{!}{\begin{tikzpicture}[x=1cm, y=1cm, thick, mid slashes/.style={decoration={markings,
						mark=at position 0.5 with {
								\draw[thick] (-3pt,-1pt) -- (0,2pt); %
								\draw[thick] (0,-1pt) -- (3pt,2pt); %
							}}, postaction={decorate}},
		mid slash/.style={decoration={markings, mark=at position 0.5 with {
								\draw[-] (-2pt,-2pt) -- (2pt,2pt);
								\draw[-] (-2pt,2pt) -- (2pt,-2pt);
							}}, postaction={decorate}},
		mid sla/.style={decoration={markings, mark=at position 0.5 with {
								\draw[-] (-2pt,-2pt) -- (2pt,2pt);
							}}, postaction={decorate}}
	]
	\node[ellipse, minimum height = 2cm, minimum width = 5.3cm, fill=colorgreenfull!10, , rotate=330] (v2) at (2,1.5) {};
	\node[ellipse, minimum height = 1cm, minimum width = 2cm, fill=colorgreenfull!20, rotate=315] (v1) at (1.5,2) {};

	\node[circle, fill=black, label=above:$\theta_0$, inner sep=1.5pt] (zero) at (-5, -1) {};
	\node[circle, fill=black, label=left:$\theta_\mathrm{init}$, inner sep=1.5pt] (init) at (0, 0) {};
	\node[circle, fill=black, label=above:$\theta^1$, inner sep=1.5pt] (rl1) at (0, 2) {};
	\node[circle, fill=black, label=below:$\theta^2$, inner sep=1.5pt] (rl2) at (4, 0) {};
	\draw[->, very thick] (init.center) -- (rl1.center);
	\draw[->, very thick] (init.center) -- (rl2.center);

	\draw[->, very thick] (zero.center) -- (rl1.center);
	\draw[->, very thick] (zero.center) -- (rl2.center);

	\pic [draw, angle eccentricity=-1, angle radius=1cm, "$\omega\approx 0^\circ$"] {right angle = rl1--zero--rl2};
	\pic [draw, angle eccentricity=-0.5, angle radius=0.5cm, "$\Omega\approx 90^\circ$"] {right angle = rl1--init--rl2};

	\draw[-, very thick, dashed, red] (rl1.center) to[bend left=25] node[pos=0.4, above, black] {$\theta_\mathrm{slerp}^{\lambda}$} (rl2.center);

	\draw[-, very thick, dashed, colorbluefull] (rl1.center) -- (rl2.center) node[midway, right, black] {$\theta_\mathrm{lerp}^{\lambda}$};	

	\path (rl1) to[bend left=25] coordinate[pos=0.377] (slerpMid) (rl2);
	\path (rl1) -- coordinate[pos=0.5] (lerpMid) (rl2);

	\draw[->, thick, dotted, colorredfull] (init.center) -- (slerpMid.center) node[midway, colorredfull, sloped, above] {$\delta_\mathrm{slerp}^{\lambda}$};
	\draw[->, thick, dotted, colorbluefull] (init.center) -- (lerpMid.center) node[midway, colorbluefull, sloped, below] {$\delta_\mathrm{lerp}^{\lambda}$};

	\path[mid sla] (rl1.center) -- (zero.center);
	\path[mid sla] (rl2.center) -- (zero.center);

	\path[mid slashes] (rl1.center) -- (init.center) node[midway, left, black] {$\delta^{1}$};
	\path[mid slashes] (rl2.center) -- (init.center) node[midway, below, black] {$\delta^{2}$};	

	\path[mid slash] (rl1.center) -- (lerpMid.center);
	\path[mid slash] (rl2.center) -- (lerpMid.center);
\end{tikzpicture}}%
	\end{center}%
	\caption{Illustration of the difference between the full weights $\theta^m$ and their task vectors $\delta^m=\theta^m - \theta_\mathrm{init}$, where darker areas are of better performance. We found in \Cref{app:slerp:angle} that $\Omega\approx90^\circ$ where $\Omega$ is the angle between task vectors such as $\cos \Omega = \frac{\delta^1 \cdot \delta_2}{\lVert \delta^1 \rVert \lVert \delta^2 \rVert}$, while $\omega$ the angle between the full weights such as $\cos \omega = \frac{\theta^1 \cdot \theta_2}{\lVert \theta^1 \rVert \lVert \theta^2 \rVert}$ satisfies $\omega\approx0^\circ$.}%
	\label{app:fig:mergeinit}%
\end{figure*}
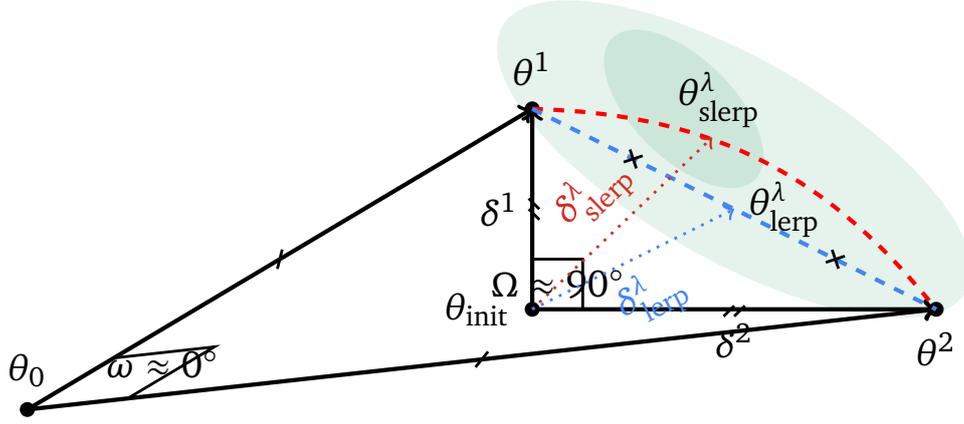

In \Cref{fig:warp_tikz}, we propose an alternative illustration of \WARP, where the different stages are more detailed than in \Cref{fig:main:warp}.
Then in \Cref{app:fig:mergeinit}, we also refine our illustration showcasing the similarity and difference between \SLERP and \LERP.
\FloatBarrier

\FloatBarrier
\FloatBarrier
\FloatBarrier
\section{Theoretical insights on task vectors, \SLERP, \LERP and \LITI}
\label{app:slerp:theoryslerpvslerp}

Based on the insights from \cite{2022arXiv221204089I} that task vectors (the differences between a fine-tuned model and its initialization) are semantically manipulable and interpretable units in the weight space, we compare \SLERP and \LERP merging operations by analyzing their task vectors.

\textbf{Background.}
Linear interpolation (\LERP)~\cite{utans1996weight} is the simplest merging strategy, notably used in the model soups variants \cite{Wortsman2022ModelSA}, and defined as:
\begin{align}
	\operatorname{lerp}\left(\theta^1, \theta^2, \lambda\right) & =(1-\lambda) \cdot \theta^1 + \lambda \cdot \theta^2.
	\label{eq:lerp}
	\tag{\LERP}
\end{align}
Then, as illustrated in \Cref{app:fig:mergeinit}, the task vector for \LERP with interpolating coefficient $\lambda$ is given by: $\delta_\mathrm{lerp}^{\lambda} = \operatorname{lerp}\left(\theta^1, \theta^2, \lambda\right) - \theta_\mathrm{init} = (1-\lambda) \cdot \delta^1 + \lambda \cdot \delta^2 $. Similarly, we define $\delta_\mathrm{slerp}^{\lambda} = \operatorname{slerp}\left(\theta_\mathrm{init}, \theta^1, \theta^2, \lambda\right) - \theta_\mathrm{init}$ where $\operatorname{slerp}$ is defined in \Cref{eq:slerp}.

\subsection{Theoretical insights on the \SLERP and \LERP task vectors}
We denote $\Omega$ the angle between the the task vectors $\delta^1$ and $\delta^2$:
\begin{equation}
	\cos \Omega = \frac{\delta^1 \cdot \delta_2}{\lVert \delta^1 \rVert \lVert \delta^2 \rVert}.
	\label{eq:omega}
\end{equation}
Based on the empirical observations from \cite{jang2024model}, confirmed in our \Cref{app:fig:task_vector_norms}, we introduce the following \Cref{assumption:normalizeddelta} for simplicity.
\begin{assumption}[Task vectors of equal norm]
	Independently fine-tuned task vectors have a same norm $l$:
	\begin{equation}
		\lVert \delta^1 \rVert = \lVert \delta^2 \rVert = l.
	\end{equation}
	\label{assumption:normalizeddelta}
\end{assumption}%
\begin{lemma}[\SLERP task vector]
	Under \Cref{assumption:normalizeddelta}, \SLERP preserves the norm of the task vector:
	\begin{equation}
		\lVert \delta_\mathrm{slerp}^{\lambda}\rVert=l.
	\end{equation}
	\label{lemma:slerpkeepsnorm}
\end{lemma}%

\begin{proof}
	By definition,
	\begin{align}
		\delta_\mathrm{slerp}^{\lambda} = \frac{\sin [(1-\lambda) \Omega]}{\sin \Omega} \cdot \delta^1+\frac{\sin [\lambda \Omega]}{\sin \Omega} \cdot \delta^2
	\end{align}
	Then, as $\delta^1 \cdot \delta^2 = l^2 \cos \Omega $,
	\begin{align}
		\frac{\lVert \delta_\mathrm{slerp}^{\lambda}\rVert^2}{l^2} & = \left(\frac{\sin [(1-\lambda) \Omega]}{\sin \Omega}\right)^2 + 2 \frac{\sin [(1-\lambda) \Omega]}{\sin \Omega} \frac{\sin [\lambda \Omega]}{\sin \Omega} \cos(\Omega) + \left(\frac{\sin [\lambda \Omega]}{\sin \Omega}\right)^2
		\\
		                                                           & = \frac{\sin^2 [(1-\lambda) \Omega] + 2 \sin [(1-\lambda) \Omega] \sin [\lambda \Omega] \cos(\Omega) + \sin^2 [\lambda \Omega]}{\sin^2 \Omega}                                                                                     \\
		                                                           & = \frac{\sin^2 \Omega}{\sin^2 \Omega}                                                                                                                                                                                              \\
		                                                           & = 1
	\end{align}
	using trigonometric identities, proving \Cref{lemma:slerpkeepsnorm}.
\end{proof}%

\begin{lemma}[\LERP task vector]
	Under \Cref{assumption:normalizeddelta}, \LERP reduces the norm of the task vector:
	\begin{equation}
		\lVert \delta_\mathrm{lerp}^{\lambda}\rVert = l \sqrt{ 1 - 2(1 - \cos\Omega)(\lambda - \lambda^2)}.
	\end{equation}
	\label{lemma:lerpreducesnorm}
\end{lemma}%

We recover that averaging weights with $\lambda=0.5$ tends to reduce the norm of the task vectors, as previously highlighted in \cite{jang2024model}.

\begin{proof}
	By definition:
	\begin{align}
		\delta_\mathrm{lerp}^{\lambda} = (1-\lambda) \cdot \delta^1 + \lambda \cdot \delta^2.
	\end{align}
	Then, as $\delta^1 \cdot \delta^2 = l^2 \cos \Omega $,
	\begin{align}
		\frac{\lVert \delta_\mathrm{slerp}^{\lambda}\rVert^2}{l^2} & = (1-\lambda)^2 + 2 \lambda (1-\lambda) \cos\Omega + \lambda^2 \\
		                                                           & = 1 - 2\lambda (1 - \cos\Omega) + 2\lambda^2 (1 - \cos\Omega)  \\
		                                                           & = 1 - 2(1 - \cos\Omega)(\lambda - \lambda^2),
	\end{align}
	proving \Cref{lemma:lerpreducesnorm} when $0<\lambda<1$.
\end{proof}

\subsection{Theoretical insights on the \KL}
\subsubsection{Linear regime}

\begin{assumption}[Linear regime \cite{Wortsman2022robust}]
	\label{assumption:linear}
	We assume that the predictions of a model $f$, with weights initialized from $\theta_0$ and fine-tuned into $\theta$, can be approximated by first-order Taylor expansion: $\forall \vx,$
	\begin{equation}
		f(\vx, \theta) \approx f(\vx, \theta_0) + \left(\theta-\theta_0\right) \cdot \nabla_{\theta} f\left(\vx, \theta_0\right).
	\end{equation}
\end{assumption}
\Cref{assumption:linear} defines a neural tangent \cite{Jacot2018} space in which the relationship between weights and functions is linear.
As previously argued in \cite{Wortsman2022ModelSA,rame2022diwa}, this Taylor expansion is reasonable partly because weights remain close during fine-tunings \cite{Neyshabur2020}, as confirmed in \Cref{app:fig:expes_stock} where they have equal norms and a cosine of one.
Yet, please note that \cite{ortiz2023task} highlighted some limitations.%
\subsubsection{\KL variations for \LERP}
\label{app:slerp:klforlerp}
We consider $\theta^1$ and $\theta^2$ weights fine-tuned from a shared SFT initialization $\theta_\mathrm{sft}$. Then in the linear regime from \Cref{assumption:linear}, weight and prediction ensembling behaves similarly:
\begin{equation}
	f\big(\vx, (1-\lambda) \cdot \theta^1 + \lambda \cdot \theta^2 \big) \approx (1-\lambda) \cdot f(\vx,  \theta^1) + \lambda \cdot f(\vx, \theta^2).
 \label{eq:linear_assumption}
\end{equation}
This similarity enables to prove the following \Cref{lemma:lerpreduceskl}.

\begin{lemma}[LERP reduces \KL]
	\label{lemma:lerpreduceskl}
	For an interpolating coefficient $0\leq\lambda\leq1$, denoting $\pi_{\lambda}$ the \LERP policy from weight interpolation $(1-\lambda) \cdot \theta^1 + \lambda \cdot \theta^2$, and $\hat{\pi}_{\lambda}$ the ensembling policy from prediction interpolation $(1-\lambda) \cdot \pi_{\theta^1} + \lambda \cdot \pi_{\theta^2}$, then under \Cref{assumption:linear},
	\begin{equation}
		\Kl(\pi_\lambda || \pi_{\theta_\mathrm{sft}}) \approx \Kl(\hat{\pi}_\lambda || \pi_{\theta_\mathrm{sft}}) \leq (1-\lambda) \Kl(\pi_{\theta^1} || \pi_{\theta_\mathrm{sft}}) + \lambda \Kl(\pi_{\theta^2} || \pi_{\theta_\mathrm{sft}}),
	\end{equation}
	\ie the \KL for \LERP is lower than the interpolated \KL.
\end{lemma}

\begin{proof}
The following proof applies the linear assumption and properties of the KL divergence.

\textbf{Approximation of KL.} The first approximate equality is a direct application of \Cref{assumption:linear} to $\pi_\lambda$. Precisely, applying \Cref{eq:linear_assumption} to the definition of $\pi_\lambda = \pi_{(1 - \lambda)\theta_1 + \lambda \theta_2}$ yields that $\pi_\lambda \approx \hat{\pi}_\lambda$.

\textbf{Upper bound of the KL.} The KL divergence is convex in both its arguments \cite{csiszar1975divergence}, thus we directly have that
\begin{equation}
    \Kl((1-\lambda) \cdot \pi_{\theta^1} + \lambda \cdot \pi_{\theta^2}|| \pi_{\theta_\mathrm{sft}}) \leq (1-\lambda) \Kl(\pi_{\theta^1} || \pi_{\theta_\mathrm{sft}}) + \lambda \Kl(\pi_{\theta^2} || \pi_{\theta_\mathrm{sft}}),
\end{equation}
which completes the proof.

\end{proof}

\begin{remark}
\Cref{lemma:lerpreduceskl} shows that the LERP $\pi_{\lambda}$ is closer in \KL to the original \SFT initialization. This relates to \Cref{lemma:lerpreducesnorm}, where we show that the linear interpolation reduces the norm to the initialization. As the interpolation brings the weights of the models closer, it is natural that it would also bring the resulting policies closer.
\end{remark}

\subsubsection{\KL and reward variation for \LITI}
\label{app:slerp:klforliti}
We now consider a given weight $\theta$ (in practice either obtained from \LERP or \SLERP of multiple fine-tuned weights) and its associated task vector $\delta$ = $\theta - \theta_\mathrm{sft}$.
In the linear regime from \Cref{assumption:linear}, for each $\eta \in [0,1]$, we have the following:
\begin{equation}
	f\left(\vx,  \theta_\mathrm{sft} + \eta \cdot \delta \right) - f\left(\vx,  \theta_\mathrm{sft}\right) \approx \eta \cdot \left( f\left(\vx, \theta_\mathrm{sft} + \delta\right) - f\left(\vx,  \theta_\mathrm{sft}\right) \right).
	\label{eq:liti-linear}
\end{equation}
We try to show that:
\begin{equation}
	\Kl\left(\pi_{\theta_\mathrm{sft} + \eta \cdot \delta} \| \pi_{\theta_\mathrm{sft}}\right) \leq \eta \cdot \Kl\left(\pi_{\theta_\mathrm{sft} + \delta} \| \pi_{\theta_\mathrm{sft}}\right).
\end{equation}

\begin{lemma}[\KL upper bound for interpolated distributions]

    For an interpolating coefficient $0\leq\eta\leq1$, denoting $\pi_{\eta}$ the \LITI policy from weight interpolation $\theta_\mathrm{sft} + \eta \cdot \delta$, and $\hat{\pi}_{\eta}$ the ensembling policy from prediction interpolation $(1- \eta) \cdot \pi_{\theta_\mathrm{sft}} + \eta \cdot \pi_{\theta_\mathrm{sft} + \delta}$, then under \Cref{assumption:linear},
  
	For each $\eta \in [0,1]$, we have that
	\begin{equation}
		\Kl\left(\pi_{\eta} \| \pi_{\theta_\mathrm{sft}}\right)
		\approx \Kl(\hat{\pi}_{\eta} \| \pi_{\theta_\mathrm{sft}}) \leq \eta \Kl\left(\pi_{\theta_\mathrm{sft} + \delta} \| \pi_{\theta_\mathrm{sft}}\right).%
	\end{equation}%
	\label{lemma:klupperbound}%
\end{lemma}

\begin{proof}
The following proof uses the same method as the one of \Cref{lemma:lerpreduceskl}. We use \Cref{assumption:linear} to link the policy with the interpolation of polices, and the inequality is a result of the KL convexity.

\textbf{Approximation of KL.} The first approximate equality is a direct application of \Cref{assumption:linear} to $\pi_\eta$. Precisely, applying \Cref{eq:liti-linear} to the definition of $\pi_\eta = \pi_{\theta_\mathrm{sft} + \eta \cdot \delta}$ yields that $\pi_\eta \approx \hat{\pi}_\eta$.

\textbf{Upper bound of the KL.} Using the fact that the KL is convex, we have
\begin{equation}
    \Kl(\eta \cdot \pi_{\theta_\mathrm{sft} + \delta} + (1- \eta) \cdot \pi_{\theta_\mathrm{sft}} || \pi_{\theta_\mathrm{sft}}) 
    \leq \eta\Kl\left(\pi_{\theta_\mathrm{sft} + \delta} \| \pi_{\theta_\mathrm{sft}}\right).
\end{equation}
\end{proof}

\begin{assumption}[LITI reward is above the expected reward]
	\label{assumption:rewardlowerbound}
	The rewards for the \LITI interpolated weights are above the interpolated rewards:
	\begin{equation}
		r(\pi_0 + \eta \cdot (\pi-\pi_{\theta_\mathrm{sft}})) \geq \eta r(\pi) + (1-\eta) r(\pi_{\theta_\mathrm{sft}}),
	\end{equation}
\end{assumption}
This \Cref{assumption:rewardlowerbound}
is based on observations from \Cref{app:fig:slerplinearmintra_controlvseta}, and extends to a reward maximization setup the notion of linear mode connectivity~\cite{Frankle2020}, usually defined \wrt the accuracy in supervised learning.

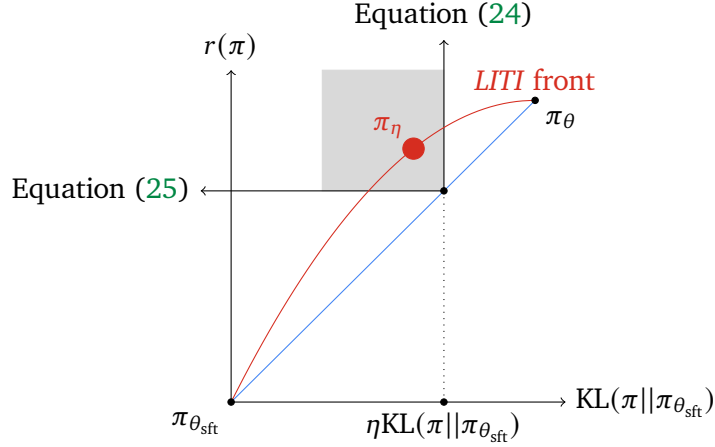
\begin{figure}
\centering
   
\begin{tikzpicture}[scale=4]

\filldraw[fill=gray!30,draw=gray!30] (0.7,0.7) rectangle (0.3,1.1); 
\draw[->] (0,0) -- (1.1,0) node[right] {$\Kl(\pi || \pi_{\theta_\mathrm{sft}})$}; 
\draw[->] (0,0) -- (0,1.1) node[above] {$r(\pi)$};
\draw[->] (0.7,0.7) -- (0.7,1.2) node[above] {\Cref{eq:reward-concave}}; 
\draw[->] (0.7,0.7) -- (-0.1,0.7) node[left] {\Cref{eq:kl-convex}};

\draw[domain=0:1,smooth,variable=\x,colorbluefull] plot ({\x},{\x});
\node[below left] at (0,0) {$\pi_{\theta_\mathrm{sft}}$};
\node[below right] at (1,1) {$\pi_{\theta}$};
\draw[domain=0:1,smooth,variable=\x,colorredfull] plot ({\x},{2*\x - \x*\x)}) node[above] {\LITI front}; 

\draw[dotted] (0.7,0.7) -- (0.7,0)  node[below] {$\eta \Kl(\pi || \pi_{\theta_\mathrm{sft}})$};;
\filldraw[black] (0.7,0.7) circle (.3pt) ;
\filldraw[black] (0,0) circle (.3pt) ;
\filldraw[black] (1,1) circle (.3pt) ;
\filldraw[black] (0.7,0.) circle (.3pt) ;
\filldraw[colorredfull] (0.6,0.84) circle (1pt) node[above left] {$\pi_\eta$};
\end{tikzpicture}
\caption{Illustration of \Cref{lemma:liti-pareto}. Based on experimental observation and theoretical insights, we see that the {\color{colorredfull} Pareto front of the \LITI policy} is better than the identity. It highlights how \Cref{eq:reward-concave,eq:kl-convex} place \LITI policies on the \KL-reward plane.\label{fig:pareto}}
\end{figure}

\begin{lemma}[LITI Pareto optimality]
\label{lemma:liti-pareto}
	Be given the convexity of the \KL from \Cref{lemma:klupperbound} and the concavity of the reward $r$ in \Cref{assumption:rewardlowerbound}, then the reward \versus \KL front of \LITI is above the diagonal. Illustration in \Cref{fig:pareto}.
\end{lemma}

\begin{proof}
	We obtain a policy $\pi_{\theta}$ fine-tuned from $\pi_{\theta_\mathrm{sft}}$. The \LITI policy for $\theta_\eta = (1 - \eta) \cdot \theta_\mathrm{sft} + \eta \cdot \theta$ is noted $\pi_\eta$. Combining the approximation from \Cref{lemma:klupperbound} and  \Cref{assumption:rewardlowerbound}, we have that
 \begin{equation}
		r(\pi_\eta) \geq (1-\eta) r(\pi_{\theta_\mathrm{sft}}) + \eta r(\pi_{\theta}).
\label{eq:reward-concave}
\end{equation}
And, from \Cref{lemma:klupperbound}, we also have that
\begin{equation}
    \Kl\left(\pi_{\eta} \| \pi_{\theta_\mathrm{sft}}\right)
		\leq \eta \Kl\left(\pi_{\theta} \| \pi_{\theta_\mathrm{sft}}\right).
\label{eq:kl-convex}
\end{equation}
This means that for every \LITI coefficient $\eta$, the \LITI policy has a higher reward than the interpolated reward at a lower KL. Geometrically, this means that each point on the Reward-KL front from LITI is on the top left quadrant of the plane according to the corresponding point on the diagonal.
\end{proof}

\subsection{Uniformly averaging $M>2$ weights with \SLERP}
\label{app:slerp:m}
The \SLERP merging formula from \Cref{eq:slerp} is only defined for $M=2$ weights.
We trivially (and certainly suboptimally) generalize this to $M>2$ weights in the uniform averaging setup, thus giving an equal coefficient to each of them, \ie $\lambda=\frac{1}{M}$.
In that setup, removing the dependency of $\theta_\mathrm{init}$ that is assumed shared, we generalize \SLERP to merge $M$ weights uniformly through the iterative procedure defined below:
\begin{equation}
	\operatorname{slerpm}\left(\{\theta^m\}_{m=1}^{M}\right) = \operatorname{slerp}\left(\operatorname{slerpm}\left(\{\theta^m\}_{m=1}^{M-1}\right), \theta^M, \lambda=\frac{1}{M}\right).
	\label{eq:slerpm}
\end{equation}
Though these operations are not associative, the standard deviations in performances are small, as indicated by the shaded areas in \Cref{fig:impact_m}.

\section{Empirical insights on task vectors, \SLERP, \LERP and \LITI}
\label{app:slerp:empirical} 
\subsection{Empirical insights on the difference between \SLERP and \LERP}
\label{app:slerp:slerpvslerplambda}

We now empirically investigate how those theoretical differences between \SLERP and \LERP affect the performance of the merged policies.

\begin{figure*}[h!]
	\begin{center}
		\begin{subfigure}[b]{0.325\textwidth}
			\includegraphics[width=0.9\textwidth]{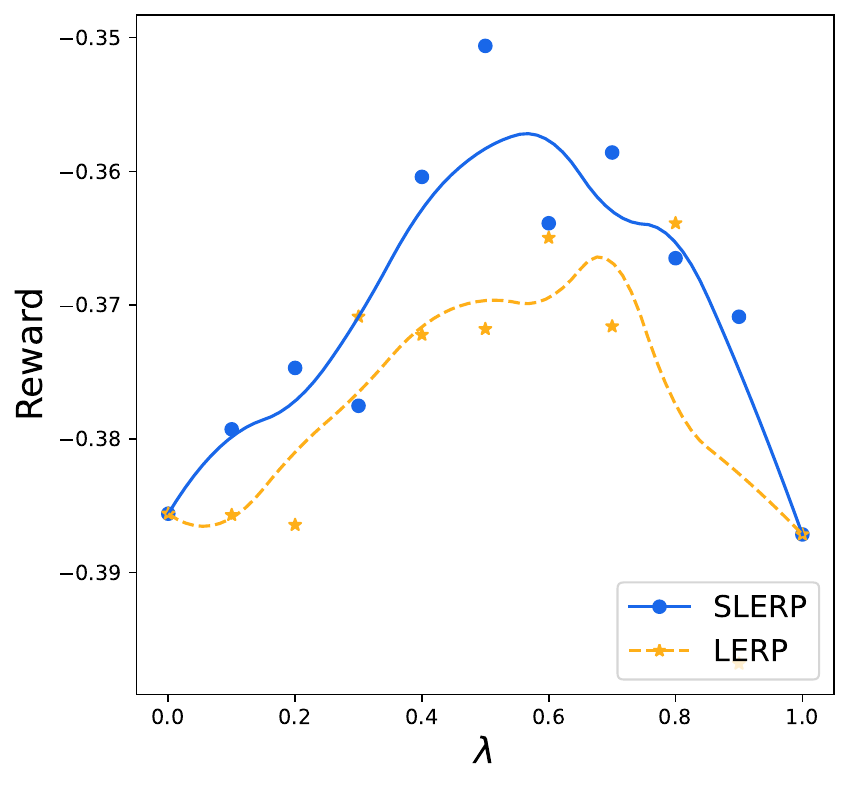}
			\caption{Reward \versus $\lambda$.}
			\label{app:fig:interpolation_controlvslambda}%
		\end{subfigure}%
		\hfill
		\begin{subfigure}[b]{0.325\textwidth}%
			\includegraphics[width=0.9\textwidth]{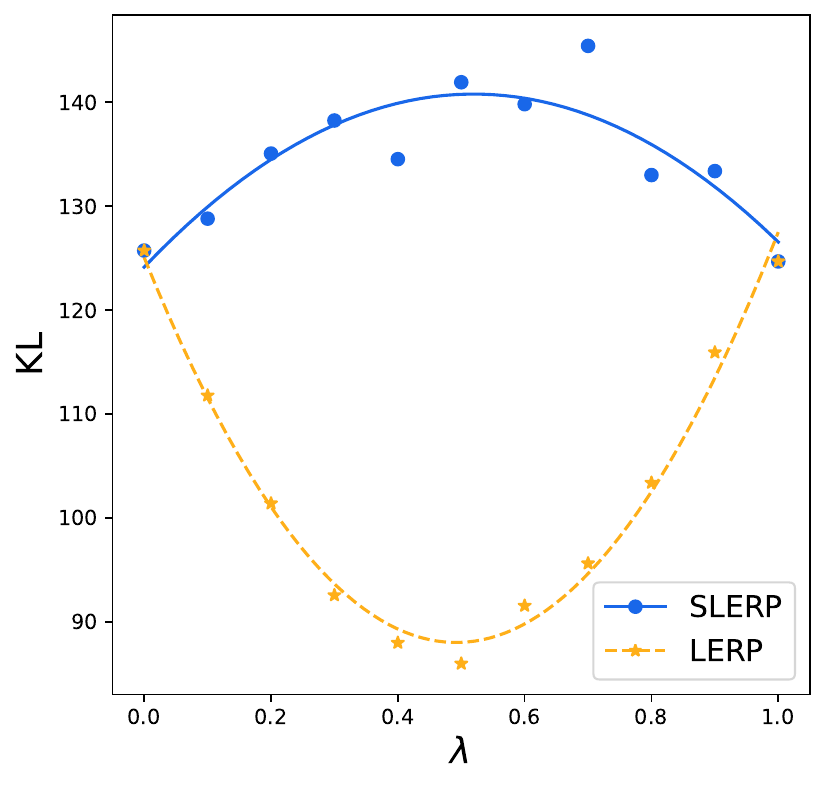}
			\caption{\KL \versus $\lambda$.}%
			\label{app:fig:klvslambda}%
		\end{subfigure}
		\hfill
		\begin{subfigure}[b]{0.325\textwidth}
			\includegraphics[width=0.9\textwidth]{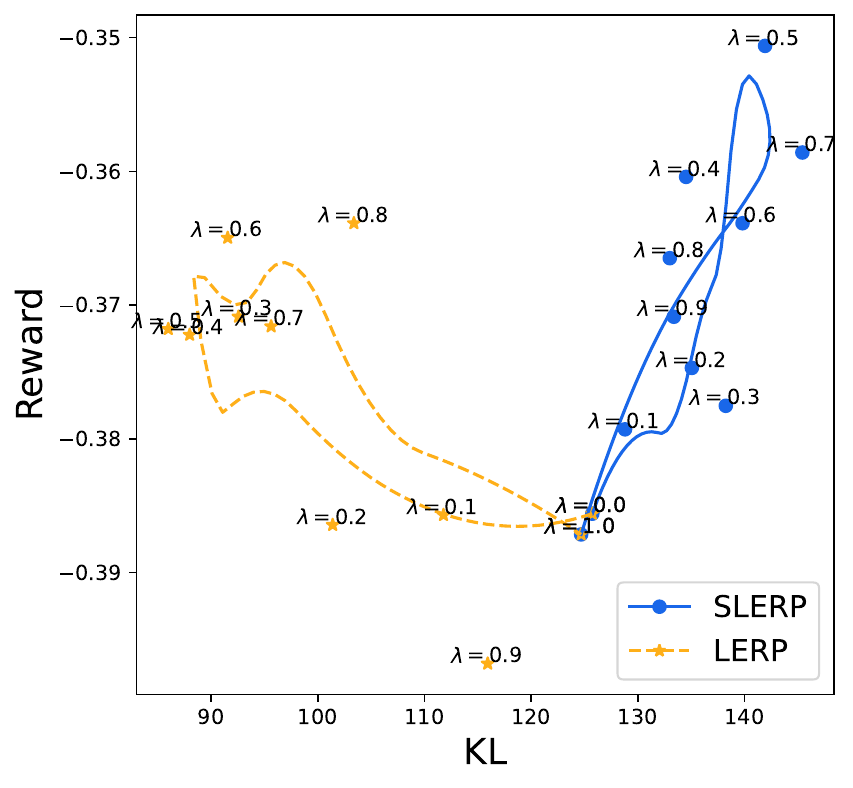}
			\caption{Reward \versus \KL.}
			\label{app:fig:interpolation_controlvskl}%
		\end{subfigure}%
	\end{center}
 \vspace{-1em} 
	\caption{\textbf{\SLERP \versus \LERP when sliding the interpolating coefficient $\lambda$}.
		Considering $M=2$ weights after $T=9k$ RL steps, we merge them using either \SLERP or \LERP, while sliding the interpolating coefficient $\lambda$ between $0$ and $1$. We then evaluate the merged checkpoints.
		\Cref{app:fig:interpolation_controlvslambda} shows that \SLERP leads to higher reward than \LERP, as previously in \Cref{fig:interpolation8090_controlvslambda}.
		\Cref{app:fig:klvslambda} shows that \LERP signicantly reduces the \KL (consistently with \Cref{lemma:lerpreduceskl}) while \SLERP slightly increases it.
		\Cref{app:fig:interpolation_controlvskl} shows how this impact the \KL-reward Pareto front, where larger markers/darker colors indicate higher values of $\lambda$; while \SLERP covers high \KL-high reward regions, \LERP tends to cover regions of lower \KL and thus also lower rewards.
		}%
	\label{app:fig:expes_slerp_lerp_lambda}%
	\begin{center}
		\begin{subfigure}[b]{0.325\textwidth}
			\includegraphics[width=0.9\textwidth]{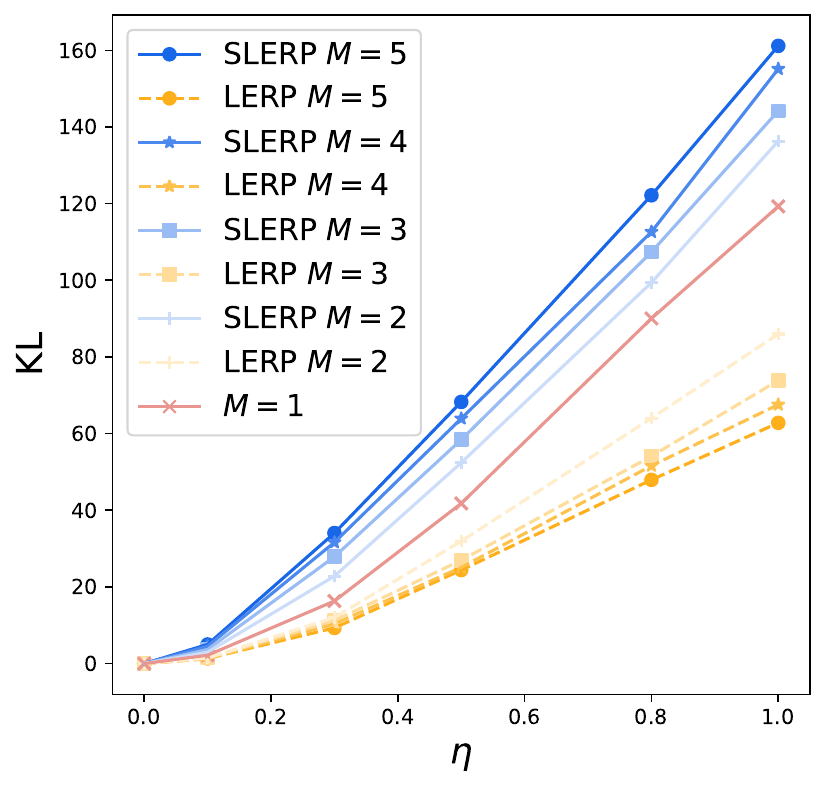}
			\caption{\KL for $\eta$.}%
			\label{app:fig:slerplinearmintra_klvseta}%
		\end{subfigure}%
		\begin{subfigure}[b]{0.325\textwidth}
			\includegraphics[width=0.9\textwidth]{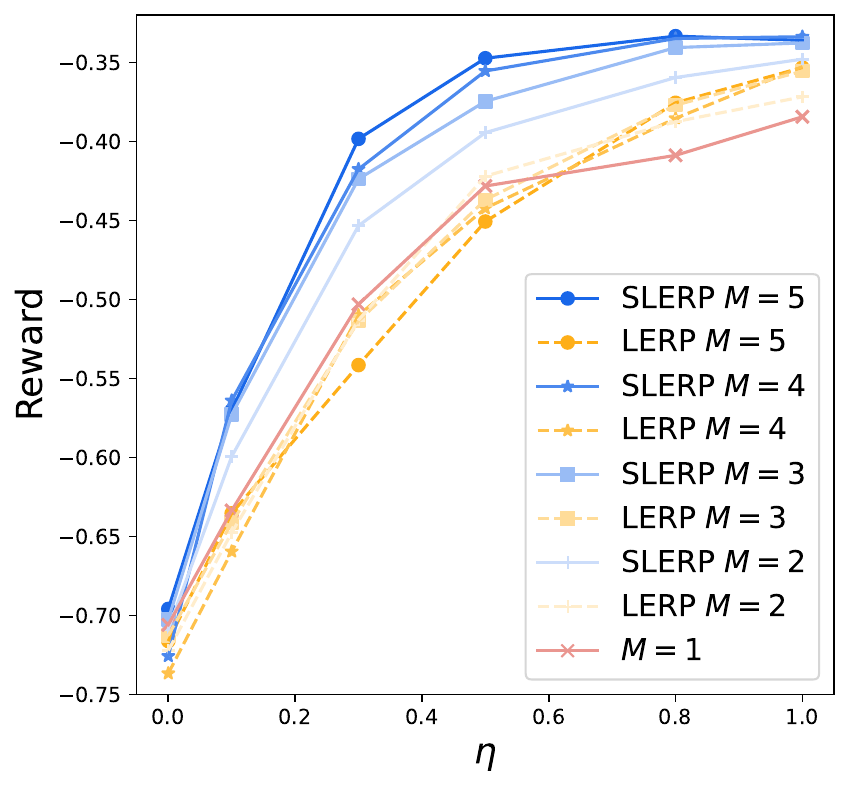}
			\caption{Reward for $\eta$.}%
			\label{app:fig:slerplinearmintra_controlvseta}%
		\end{subfigure}%
	\end{center}%
 \vspace{-1em} 
	\caption{\textbf{\SLERP \versus \LERP when sliding the interpolating coefficient $\eta$ of \LITI}.
	In \Cref{app:fig:slerplinearmintra_klvseta} we show that the \KL is convex (and almost linear) with regard to $\eta$, consistently with \Cref{lemma:klupperbound}.
	In contrast, \Cref{app:fig:slerplinearmintra_controlvseta} shows that the reward is concave, validating \Cref{assumption:rewardlowerbound}.}%
	\label{app:fig:expes_slerp_lerp_wrteta}%
	\begin{center}	
		\begin{subfigure}[b]{0.325\textwidth}
			\includegraphics[width=0.9\textwidth]{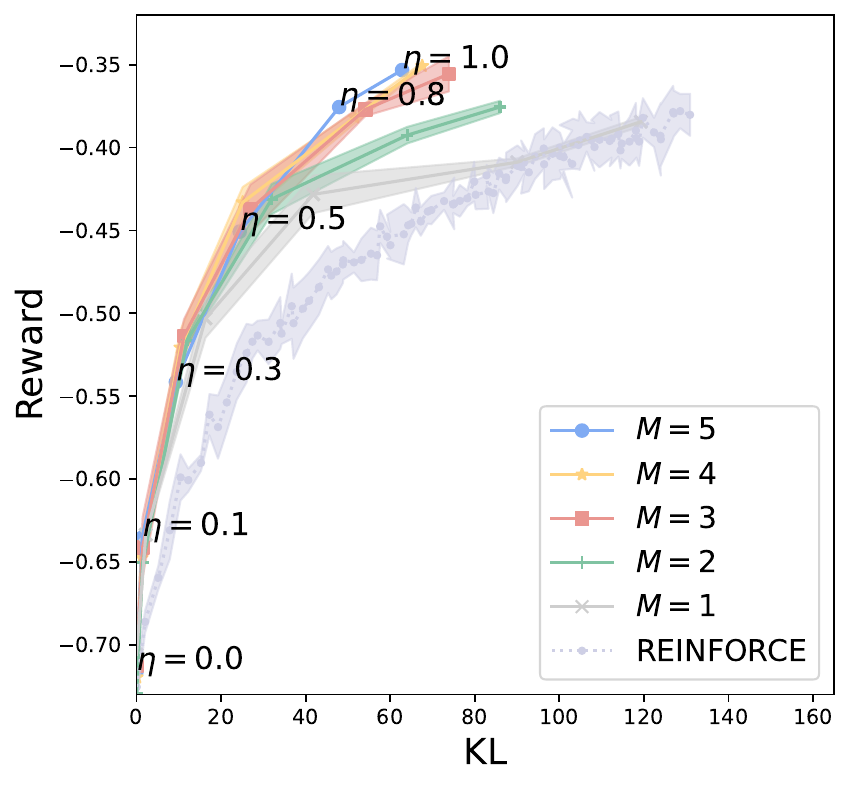}
			\caption{\LITI of \LERP of $M$ weights.}%
			\label{app:fig:linearm_controlvskl_5}%
		\end{subfigure}%
		\hfill
		\begin{subfigure}[b]{0.325\textwidth}
			\includegraphics[width=0.9\textwidth]{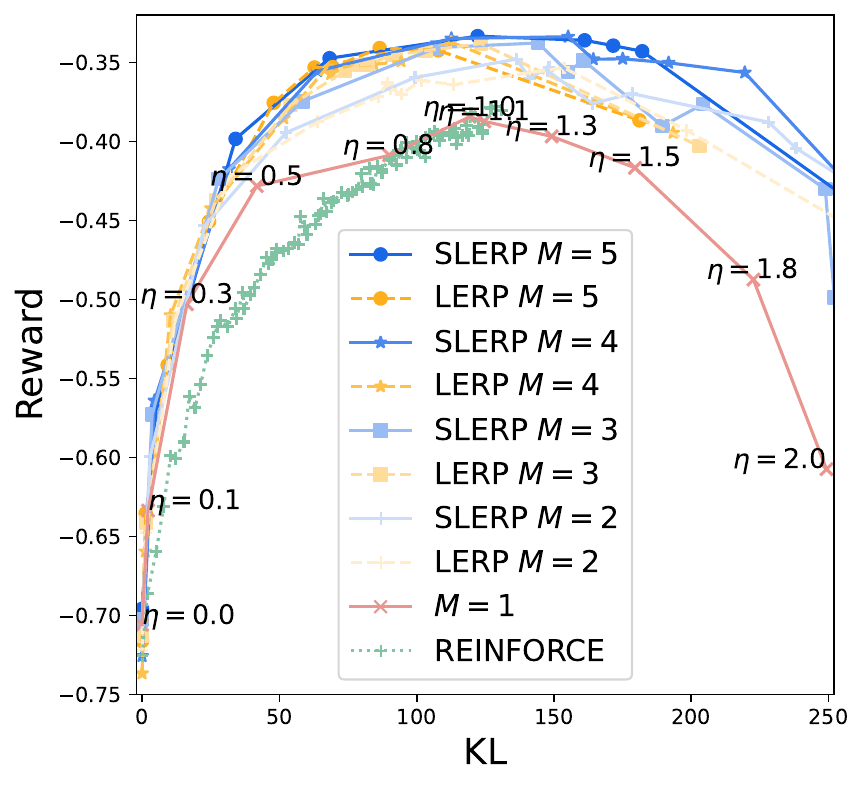}
			\caption{Extrapolation with $0\leq\eta\leq2$.}
			\label{app:fig:slerplinearm_controlvskl}%
		\end{subfigure}%
		\hfill
		\begin{subfigure}[b]{0.325\textwidth}%
			\includegraphics[width=0.9\textwidth]{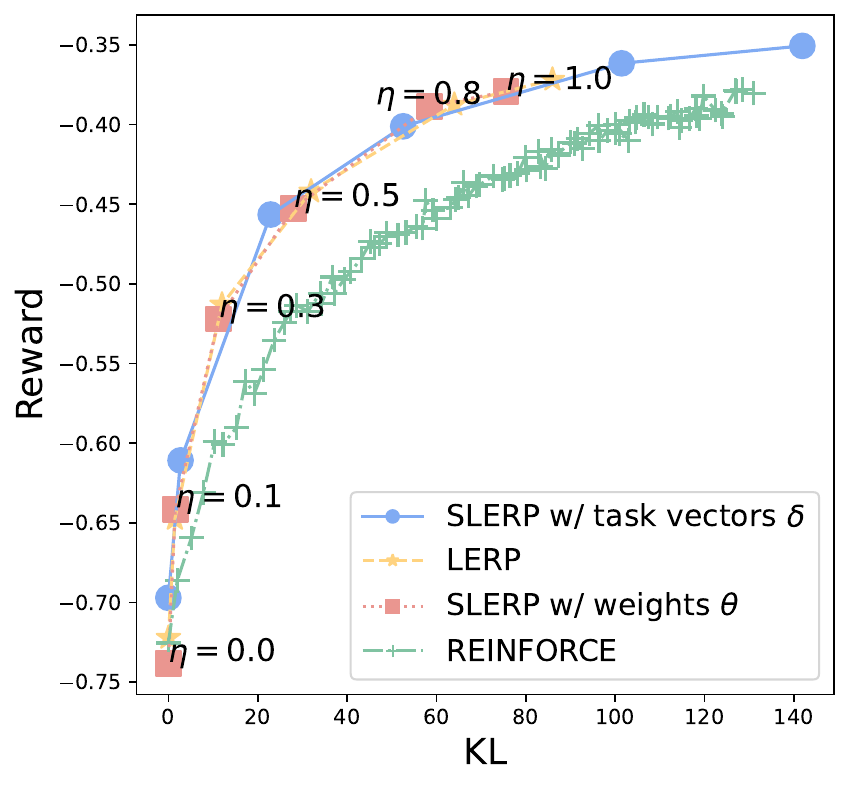}
			\caption{\SLERP w/o task vectors.}%
			\label{app:fig:slerpnotask_controlvskl}%
		\end{subfigure}
	\end{center}%
 \vspace{-1em}
	\caption{\textbf{\SLERP \versus \LERP when sliding the interpolating coefficient $\eta$ of \LITI}.
        \Cref{app:fig:linearm_controlvskl_5} merges $M$ policies with \LERP and $\lambda=\frac{1}{M}$(the endpoints on the top right of the solid lines), and then interpolates towards their \SFT init, where light-colored areas show standard deviations across $5$ experiments, and with $0\leq\eta\leq1$. In contrast, in \Cref{app:fig:slerplinearm_controlvskl} we investigate extrapolation \cite{zheng2024weaktostrong}, using $0\leq\eta\leq2$ enabling to compare the full fronts of solutions with \LERP and \SLERP. Finally, \Cref{app:fig:slerpnotask_controlvskl} confirms that applying \SLERP on the full weights $\theta$ rather than on the task vectors $\delta$ perform very similarly to \LERP.}%
	\label{app:fig:expes_slerp_lerp}%
\end{figure*}

\textbf{\SLERP \versus \LERP.}
In \Cref{app:fig:expes_slerp_lerp_lambda} we adjust the interpolating coefficient $\lambda$, highlighting distinct behaviors for \SLERP and \LERP.
\SLERP consistently enhances rewards more than \LERP, as depicted in \Cref{fig:interpolation8090_controlvslambda,app:fig:interpolation_controlvslambda}. However, a comprehensive evaluation must consider both \KL and reward. 
As shown in \Cref{app:fig:klvslambda}, \LERP consistently reduces \KL, corroborating with \Cref{lemma:lerpreducesnorm} that \LERP reduces the norm of updates (while \SLERP preserves it).
When plotting these metrics together in \Cref{app:fig:interpolation_controlvskl}, we observe that \SLERP and \LERP target different regions on the Pareto front: \SLERP achieves higher rewards at the expense of increased \KL, while the main impact of \LERP is to lower \KL.
This is consistent with \Cref{lemma:lerpreducesnorm,lemma:lerpreduceskl}, be given the orthogonal angles between task vectors $\Omega\approx90^\circ$ (as shown in \Cref{app:fig:task_vector_cosines}).

\textbf{Combining \SLERP and \LERP with \LITI.}
We also compare the behaviours of \SLERP and \LERP when we apply \LITI, as we adjust the interpolating coefficient $\eta$.
\Cref{app:fig:slerplinearmintra_klvseta} and \Cref{app:fig:slerplinearmintra_controlvseta} validate that \KL is convexe with regard to $\eta$ while the reward is concave with regard to $\eta$, for different values of $M$.
This is also highlighted in \Cref{app:fig:linearm_controlvskl_5}, which reproduces the results from \Cref{fig:impact_m} (and maintaining the same axis limits), replacing \SLERP by \LERP: this leads to critical changes in the Pareto fronts. Inded, increasing $M$ now tends to decrease \KL for \LERP, while it used to increase reward with \SLERP.
In \Cref{app:fig:slerplinearm_controlvskl}, we explore the extrapolation strategies from \cite{zheng2024weaktostrong}, using $0 \leq \eta \leq 2$ to compare the full extrapolated fronts from \LERP and \SLERP. While both perform similarly on low \KL, our results suggest that \SLERP perform better in high \KL regions.

\textbf{Conclusion.}
\SLERP demonstrates some key advantages. In particular, it reveals the full Pareto front of solutions, while \LERP only exposes a portion; extrapolation \Cref{app:fig:slerplinearm_controlvskl} with $\eta>1$ can partially mitigate this but as our experiments suggest, \LERP curves consistently lag behind \SLERP curves in high-reward regions.
Moreover, from a practical perspective, \SLERP scales the choice of $\eta$ effectively, where $1$ represents full updates and a fixed value of $0.3$ always corresponds to the same operational region, optimizing for high reward and \KL.
\FloatBarrier
\subsection{Empirical insights on the role of task vectors}
\label{app:slerp:angle}
\begin{figure*}[h!]
	\begin{center}
		\begin{subfigure}[b]{0.24\textwidth}%
			\includegraphics[width=\textwidth]{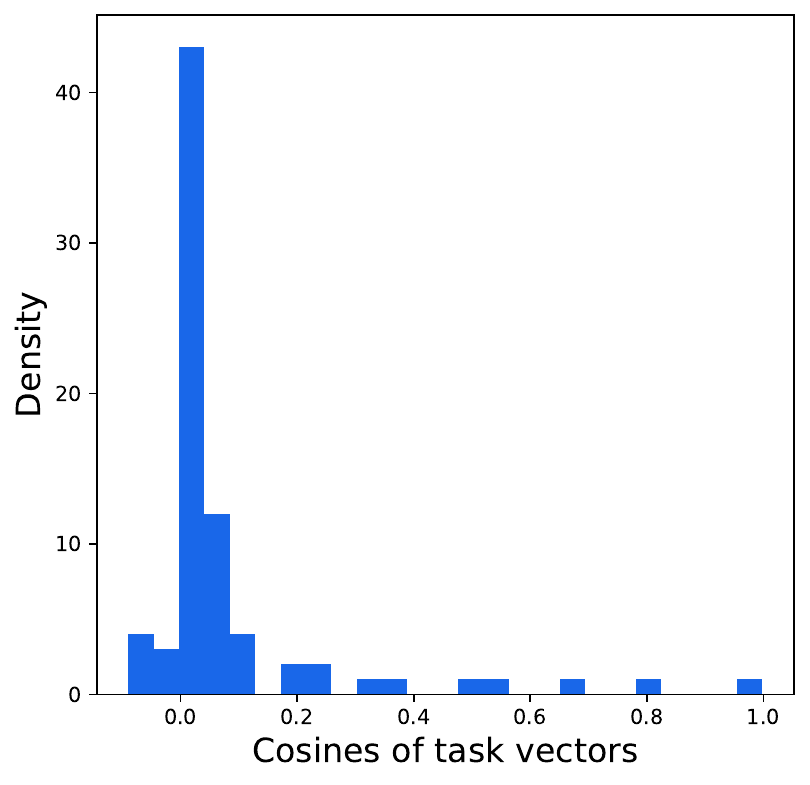}
			\caption{$\frac{\delta^1 \cdot \delta^2}{\lVert \delta^1 \rVert \lVert \delta^2 \rVert}$.}%
			\label{app:fig:task_vector_cosines}%
		\end{subfigure}
		\hfill
		\begin{subfigure}[b]{0.24\textwidth}%
			\includegraphics[width=\textwidth]{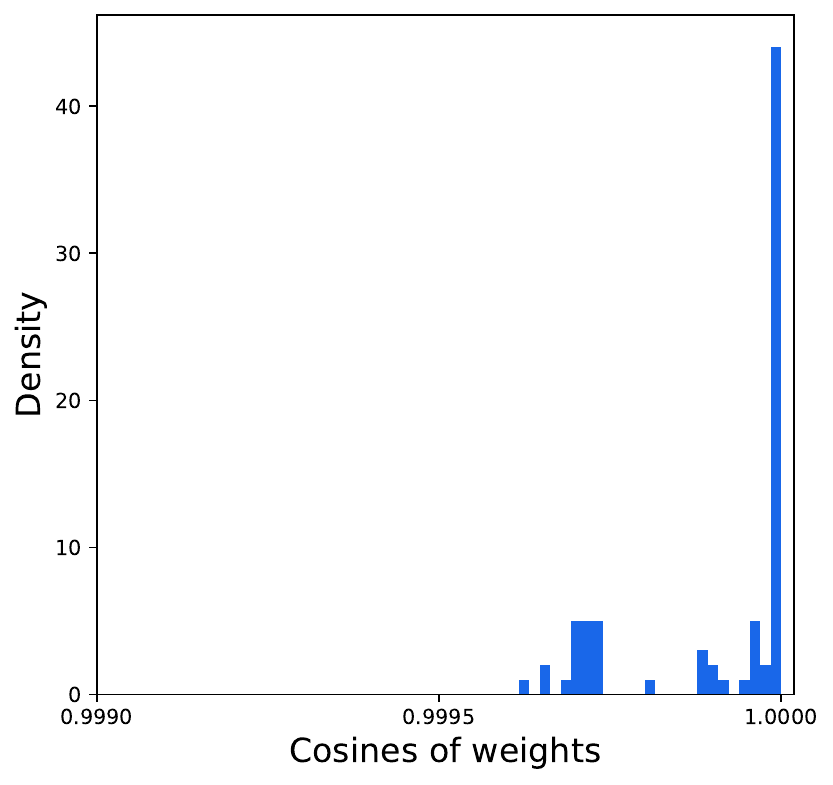}
			\caption{$\frac{\theta^1 \cdot \theta^2}{\lVert \theta^1 \rVert \lVert \theta^2 \rVert}$.}%
			\label{app:fig:weights_cosines}%
		\end{subfigure}
		\hfill
		\begin{subfigure}[b]{0.24\textwidth}%
			\includegraphics[width=\textwidth]{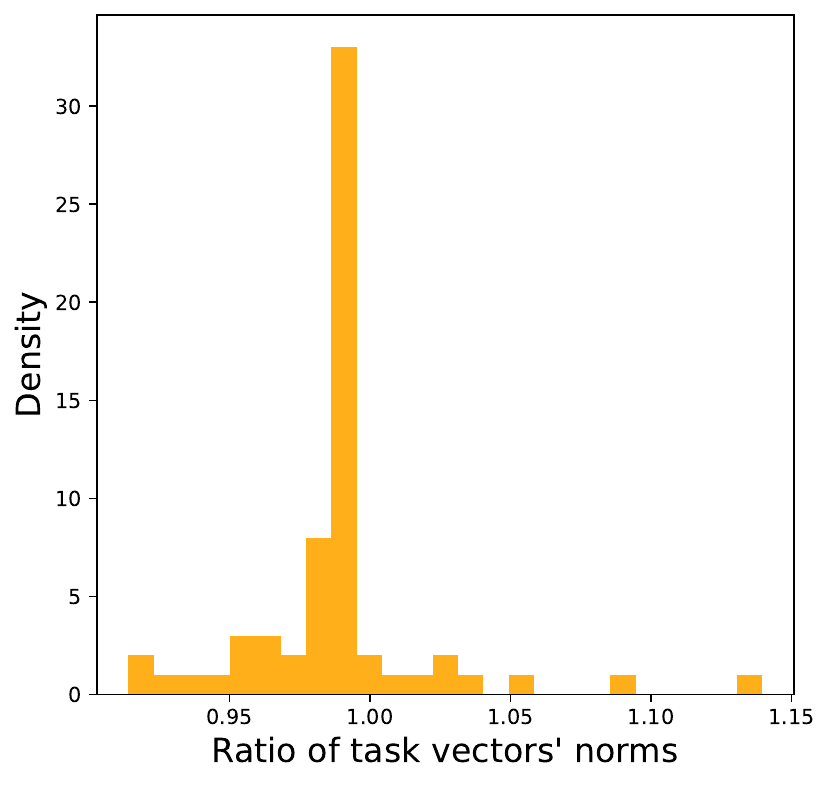}
			\caption{$\frac{\lVert \delta^1 \rVert}{\lVert \delta^2 \rVert}$.}%
			\label{app:fig:task_vector_norms}%
		\end{subfigure}
		\hfill
		\begin{subfigure}[b]{0.24\textwidth}%
			\includegraphics[width=\textwidth]{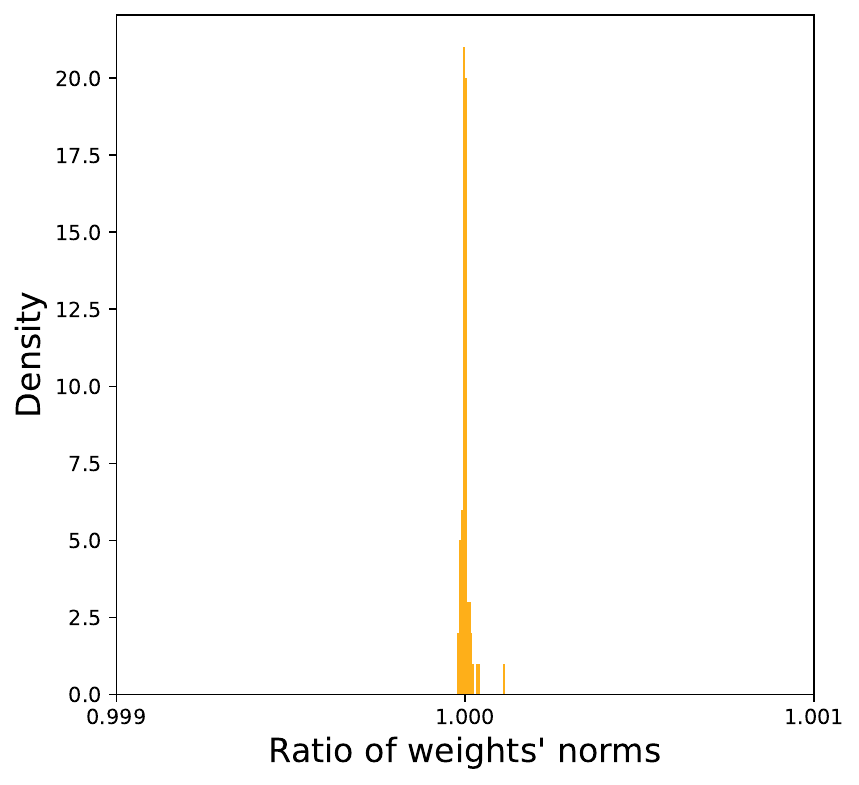}
			\caption{$\frac{\lVert \theta^1 \rVert}{\lVert \theta^2 \rVert}$.}%
			\label{app:fig:weight_norms}%
		\end{subfigure}
	\end{center}%
	\caption{\textbf{Angles and norms of (full) weights $\theta^m$ and their task vectors $\delta^m=\theta^m-\theta_\mathrm{init}$}. The histograms represent data across the $28$ layers of the \GEMMAB architecture. \Cref{app:fig:task_vector_cosines} plots the histograms of task vector cosines. \Cref{app:fig:weights_cosines} plots the histograms of weights cosines. \Cref{app:fig:task_vector_norms} plots the histograms of task vector norms ratio. \Cref{app:fig:weight_norms} plots the histograms of weights norms ratio.}%
	\label{app:fig:expes_stock}%
\end{figure*}

We now explore the effectiveness of applying \SLERP on task vectors $\delta$ \versus full weights $\theta$, as illustrated in \Cref{app:fig:mergeinit}.
To this end, in \Cref{app:fig:expes_stock} we draw inspiration from \cite{jang2024model} and plot the angles $\Omega$ and $\omega$ and norms of $\delta$ and $\theta$.

\textbf{Angles of task vectors $\Omega\approx90^\circ$.}
\Cref{app:fig:task_vector_cosines} shows that the task vectors are typically orthogonal ($\Omega \approx 90^\circ$), highlighting the diverse trajectories of the different RL fine-tunings.
This is in contrast with \cite{jang2024model} for supervised fine-tunings, where $\Omega$ typically range between $40^\circ$ and $80^\circ$.
We suspect that this is related to the underlying differences between reinforcement and supervised learning; in RL the policies are trained on their own generations, creating more orthogonal task vectors, whereas in supervised learning the LLM try to imitate the groundtruth labels, leading to more similar task vectors.
The orthogonality of our task vectors prevents the use of the update rule $\eta \rightarrow \frac{2 \cos \Omega}{1+\cos \Omega}$ suggested from Eq. 2 in \cite{jang2024model}, as it would lead to $\eta \approx 0$, deleting any potential update.

\textbf{Angles of full weights $\omega\approx0^\circ$.}
In contrast, \Cref{app:fig:weights_cosines} show that full weights remain collinear ($\omega\approx0^\circ$).
This explains the empirical results from \Cref{app:fig:slerpnotask_controlvskl}, where applying \SLERP directly to full weights results in behaviors similar to \LERP.
Indeed, as the angles $\omega\approx0^\circ$, spherical interpolation effect is minimal because $\sin(x)\approx x + \mathcal{O}(x^3)$, and thus $\frac{\sin [\lambda \omega]}{\sin \omega} \approx \frac{\lambda \omega}{\omega} \approx \lambda$.

\textbf{Norms consistency.}
\Cref{app:fig:task_vector_norms} confirms the consistency in the norms of different task vectors, supporting our \Cref{assumption:normalizeddelta}. This uniformity is aligned with previous research \cite{jang2024model}.
As a side note, this consistency extends to full weights $\theta$, confirming that fine-tuning typically results in minimal changes to the overall weight \cite{Neyshabur2020}.%

\FloatBarrier
\section{Empirical investigation of several design choices}
\label{app:expe}
We include several experiments showcasing the robustness of \WARP to different design choices, while further demonstrating its superiority in terms of \KL-reward Pareto optimality.
Specifically,
\Cref{app:expe:trainingsteps} analyzes the performances along training at different steps $T$;
\Cref{app:expe:ablationmubeta} provides results with different values for the hyperparameters $\mu$ and $\beta$;
\Cref{app:expe:ablationeta} shows the impact of the update rate $\eta$ to provide an improved initialization for the \nth{2} iteration of \WARP;
finally, \Cref{app:expe:ablationinit} shows that in iterative \WARP, interpolating towards the episode initialization or the \SFT initialization both perform similarly.
\FloatBarrier
\subsection{Analyzing the number of training steps}
\label{app:expe:trainingsteps}
\begin{figure*}[h!]
	\begin{center}
		\begin{subfigure}[b]{0.495\textwidth}%
			\includegraphics[width=\textwidth]{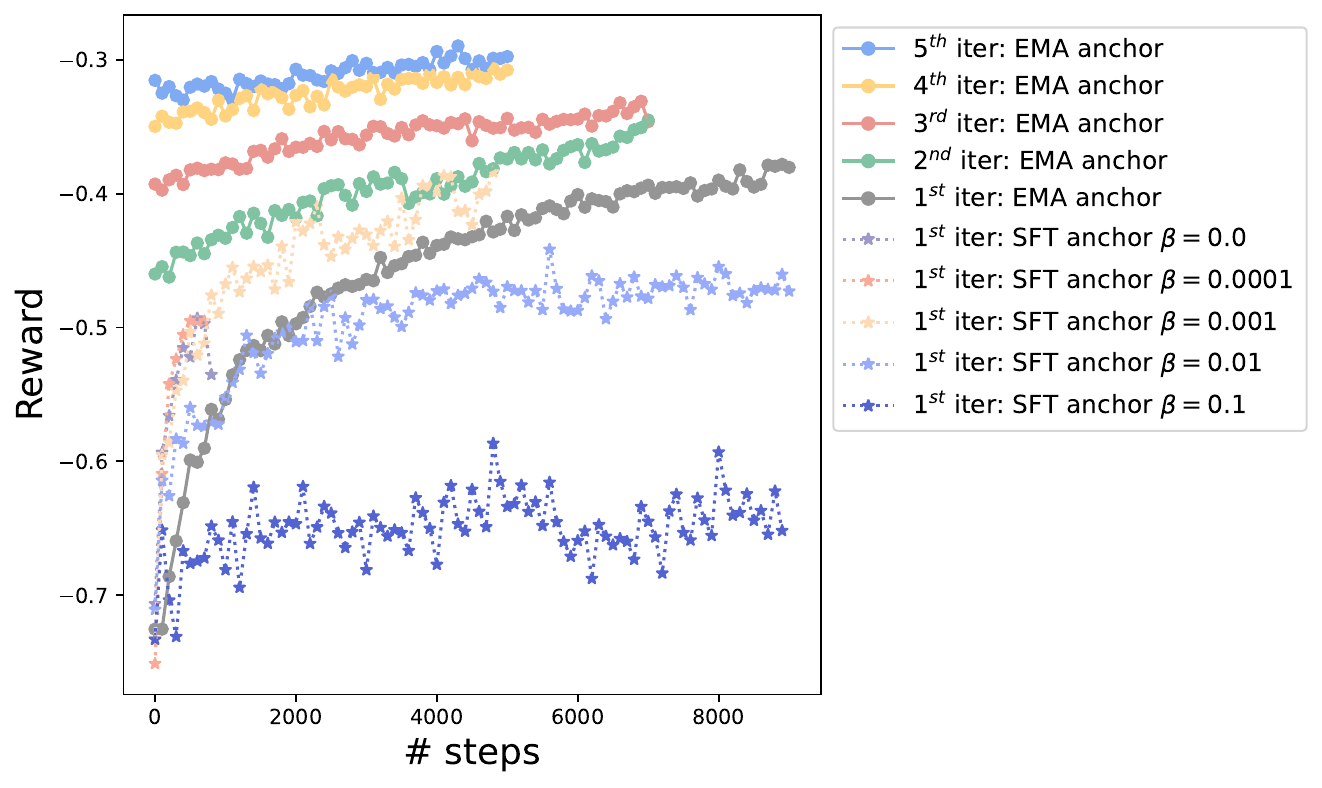}
			\caption{Reward along training.}%
			\label{app:fig:rewardvssteps_controlvsstep}
		\end{subfigure}
		\hfill
		\begin{subfigure}[b]{0.495\textwidth}%
			\includegraphics[width=\textwidth]{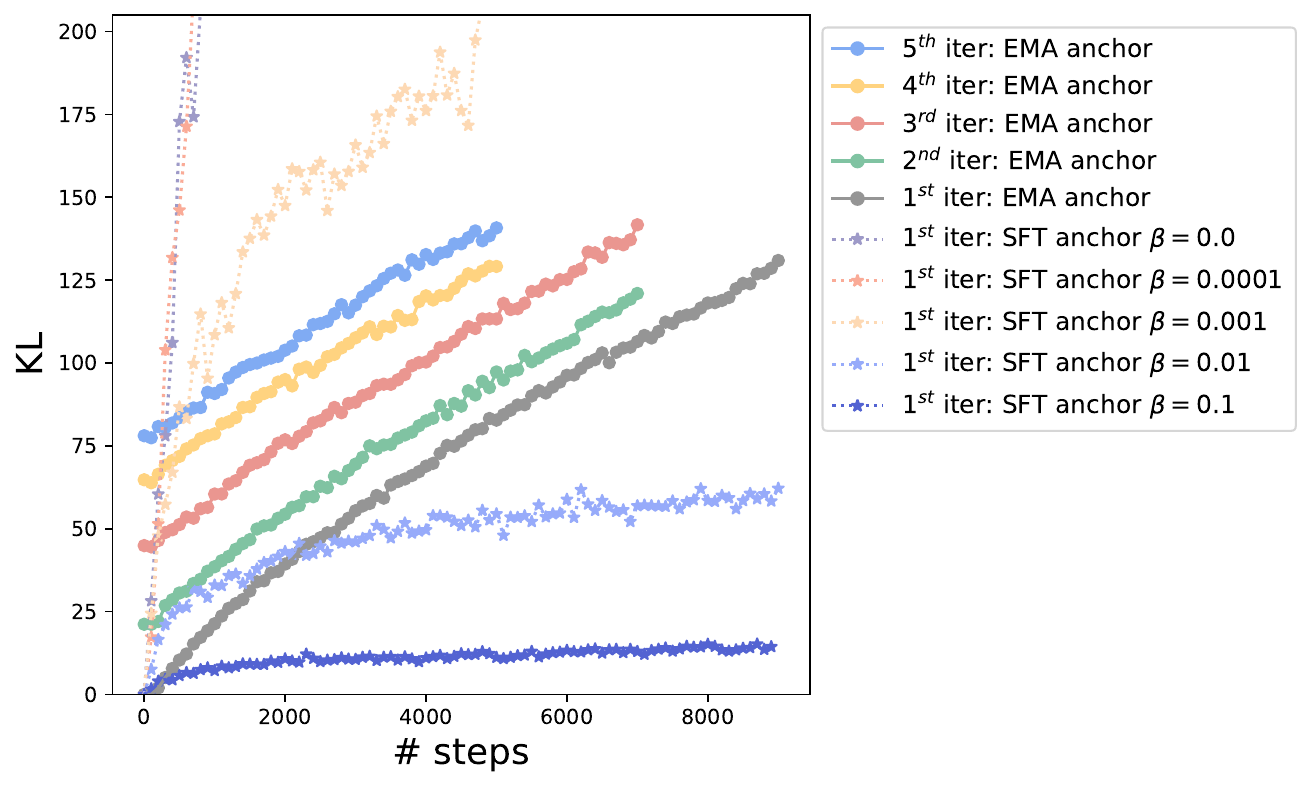}
			\caption{\KL along training.}%
			\label{app:fig:rewardvssteps_klvsstep}
		\end{subfigure}
	\end{center}%
	\caption{
		\textbf{Rewards and \KL at different number of training steps $T$}.
		\Cref{app:fig:rewardvssteps_controlvsstep,app:fig:rewardvssteps_klvsstep} complement \Cref{fig:anchor_ema_vs_sft} and \Cref{fig:warp_vs_iterations}, this time plotting rewards and \KL separately as a function of the number of training steps $T$.
		Regarding iterative \WARP, we observe that each iteration has higher rewards but also higher \KL (by starting at training step $0$ from a new initialization).
		Regarding the baseline (REINFORCE with SFT anchor), we observe that low values of $\beta$ lead to very fast hacking of the reward, as visible by the \KL exploding, while high values of $\beta$ slow down the learning procedure.
	}%
	\label{app:fig:perfsalongtrainingsteps}%
\end{figure*}%
\FloatBarrier
\begin{figure*}[h!]
	\begin{center}	
	\begin{subfigure}[b]{0.325\textwidth}%
		\includegraphics[width=\textwidth]{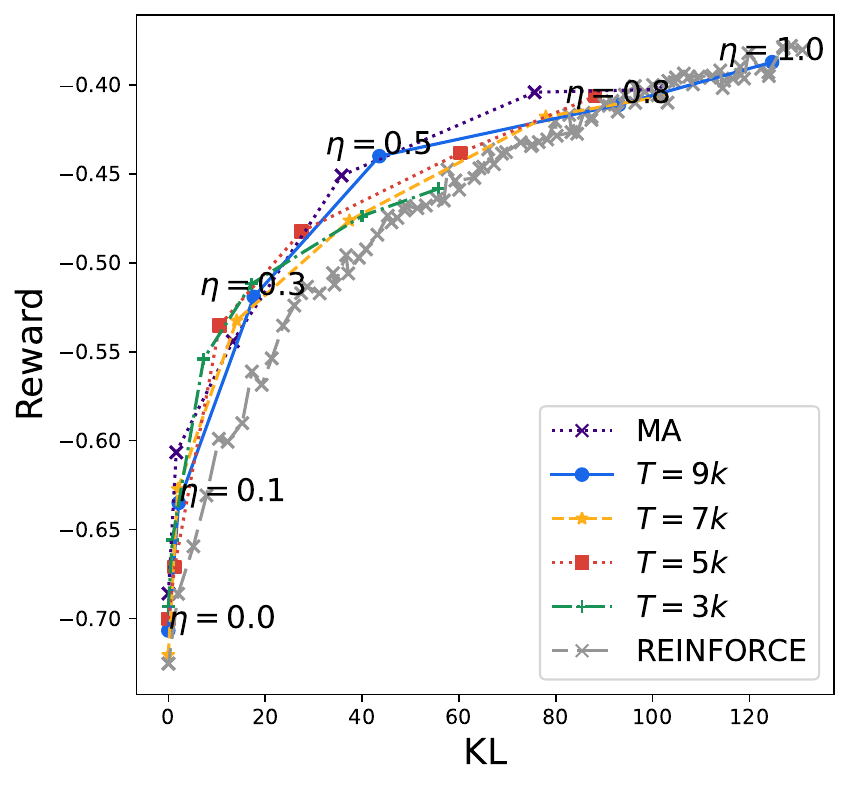}
	\end{subfigure}
	\end{center}%
	\caption{
		\textbf{\LITI with $M=1$ at different number of training steps $T$}.
		The reward gain is significantly reduced compared to \Cref{fig:num_steps} where we first merged $M=2$ policies before applying \LITI.
		We also try to perform moving average (MA)~\cite{izmailov2018,arpit2021ensemble} before applying \LITI, averaging multiple checkpoints collected along a single RL fine-tuning at steps $\{6k, 7k, 8k, 9k\}$; this does not improve performances, suggesting the need to merge weights from independent fine-tunings to have enough diversity \cite{rame2022diwa}.
	}%
	\label{app:fig:wiseatstep_controlvskl}%
\end{figure*}%
\FloatBarrier
\clearpage
\subsection{Analyzing the values of $\mu$ and $\beta$}
\label{app:expe:ablationmubeta}
\FloatBarrier
\begin{figure*}[h!]
	\begin{center}
		\begin{subfigure}[b]{0.325\textwidth}
			\includegraphics[width=\textwidth]{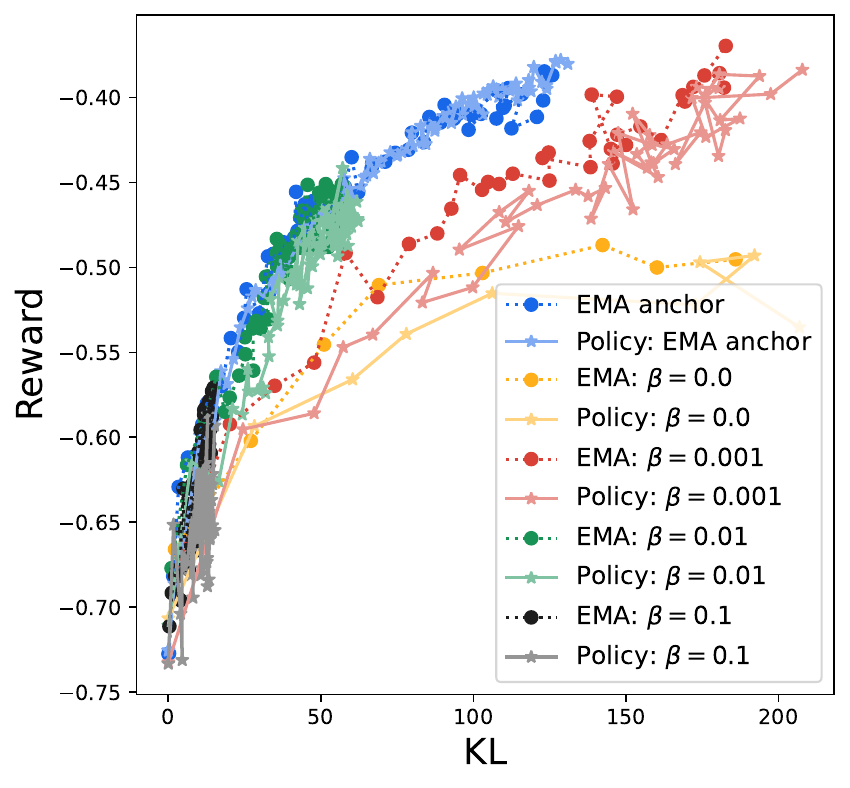}
			\caption{Reward \versus \KL.}
			\label{fig:emateacher_vs_student_rewardkl}%
		\end{subfigure}%
		\hfill		
		\begin{subfigure}[b]{0.325\textwidth}
			\includegraphics[width=\textwidth]{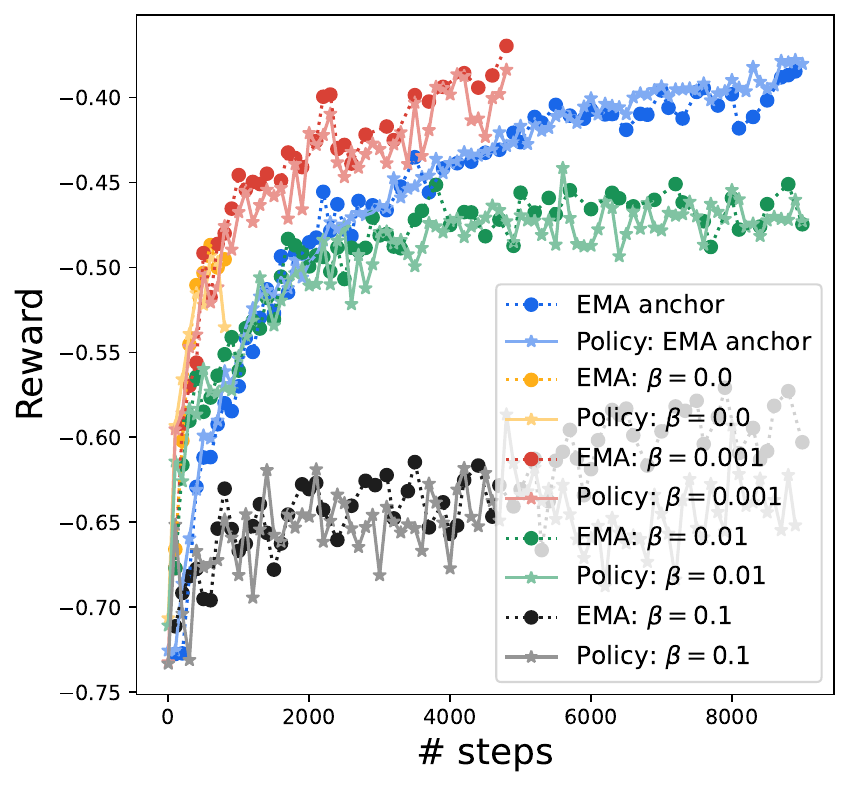}
			\caption{Reward \versus steps.}
			\label{fig:emateacher_vs_student_rewardstep}%
		\end{subfigure}%
		\hfill	
		\begin{subfigure}[b]{0.325\textwidth}
			\includegraphics[width=\textwidth]{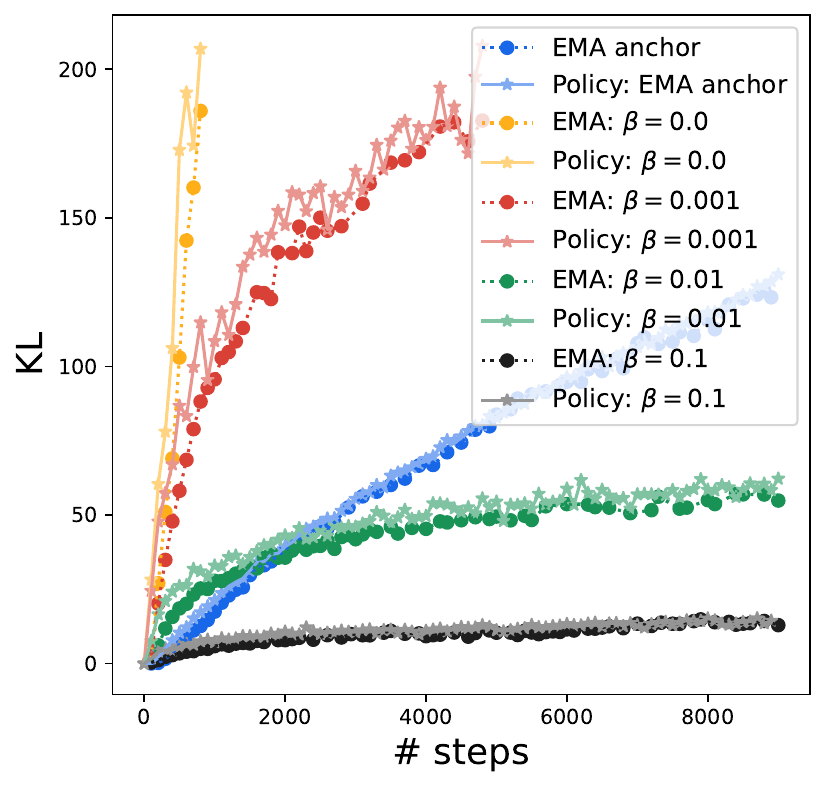}
			\caption{\KL \versus steps.}
			\label{fig:emateacher_vs_student_klstep}%
		\end{subfigure}%
		\hfill						
	\end{center}%
	\caption{\textbf{\EMA \versus their base policies}, complementing the results from \Cref{fig:contremalp_controlvsstep,fig:anchor_ema_vs_sft}.
	We confirm in \Cref{fig:emateacher_vs_student_rewardkl} that the \EMA of all variants (with SFT anchor) perform similarly or better than their base policies in terms of \KL-reward Pareto optimality.
	As a reminder, we perform evaluation every $100$ steps, and train them for $T=9k$ steps, though we stopped the trainings if the base policy ever reaches a \KL of $200$.
	This confirms \Cref{obs:ema}; the benefits of our variant with \EMA anchor is partly explained by distillation from an improved mean teacher \cite{tarvainen2017mean}.
	}
	\label{fig:emateacher_vs_student}%
\end{figure*}%
\FloatBarrier
\begin{figure*}[h!]
	\begin{center}
		\begin{subfigure}[b]{0.325\textwidth}%
			\includegraphics[width=\textwidth]{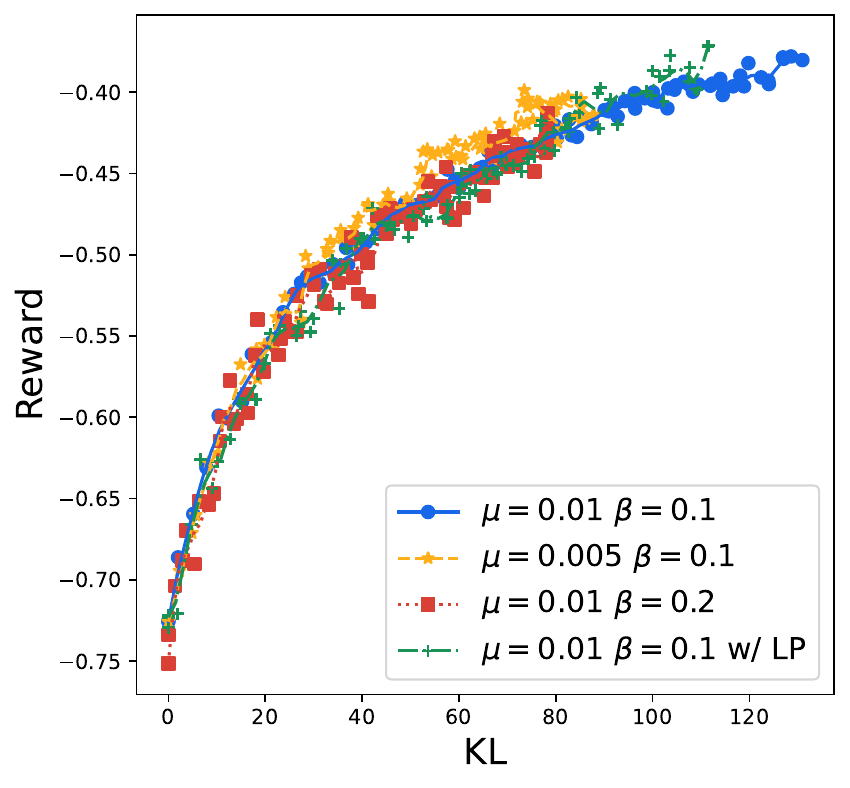}
			\caption{Reward \versus \KL.}%
			\label{app:fig:RLablation_controlvskl}%
		\end{subfigure}
		\begin{subfigure}[b]{0.325\textwidth}%
			\includegraphics[width=\textwidth]{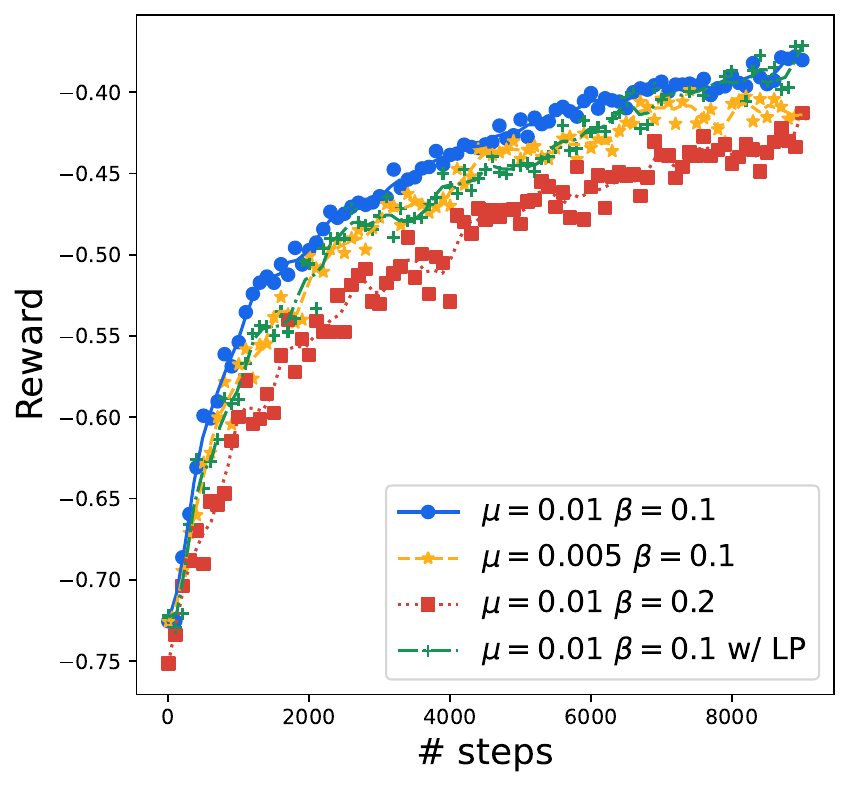}
			\caption{Reward \versus steps.}%
			\label{app:fig:RLablation_controlvsstep}%
		\end{subfigure}
	\end{center}%
	\caption{\textbf{Experiments ablating the values for the \EMA update rate $\mu$ and the \KL regularization strength $\beta$}.
		So far we have systematically used $\mu=0.01$ and $\beta=0.1$ for all \EMA-based runs, including in the iterative \WARP. These hyperparameters were chosen at the project's onset and have remained unchanged.
		In \Cref{app:fig:RLablation_controlvskl,app:fig:RLablation_controlvsstep} we increase regularization with $\mu=0.005$ and $\beta=0.2$. Our results indicate that reducing $\mu$ or increasing $\beta$ behaves similarly, marginally improving the \KL-reward Pareto front but slowing down training.
		Additionally, we include the training trajectory when using a length penalty (LP), as detailed in \Cref{app:length_penalty}.}
	\label{app:fig:expes_ablation}%
\end{figure*}%
\FloatBarrier
\clearpage
\subsection{Analyzing the values of $\eta$}
\label{app:expe:ablationeta}
\FloatBarrier
\begin{figure*}[h!]
	\begin{center}
		\begin{subfigure}[b]{0.325\textwidth}%
			\includegraphics[width=\textwidth]{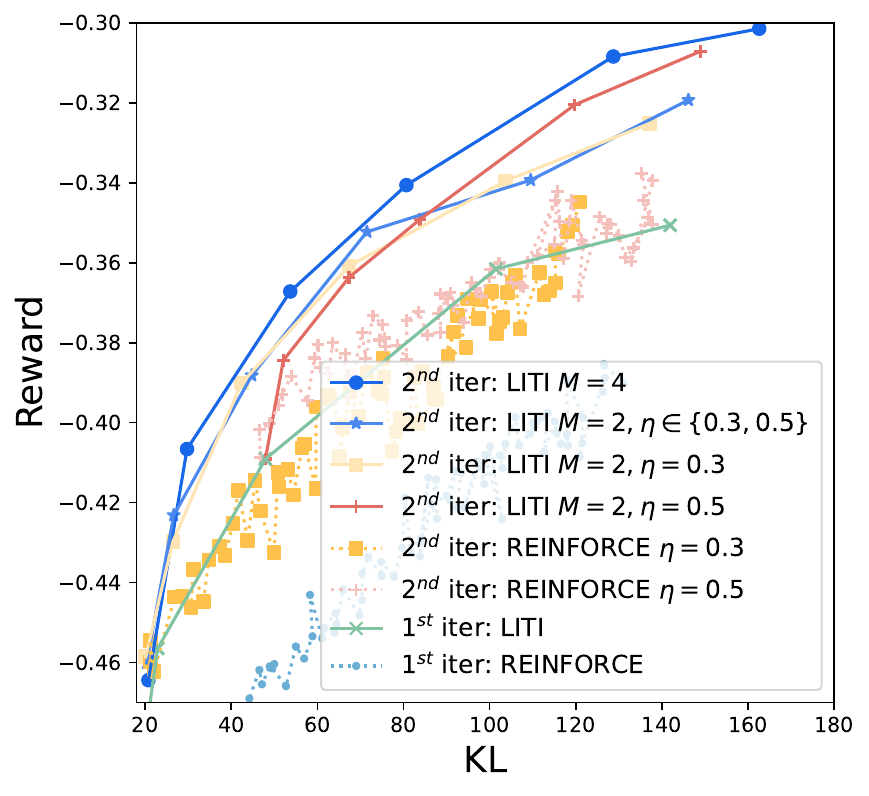}
		\end{subfigure}
	\end{center}%
	\caption{\textbf{Experiments ablating the \LITI update rate $\eta$}.
		As we initiate the \nth{2} iteration of \WARP, selecting an appropriate value for $\eta$ is key, as it determines the starting point $\theta^{\eta}$ and functions similarly to an outer learning rate (see \Cref{sec:discussion}). We usually set $\eta=0.3$, but now provide results with an increased $\eta=0.5$, starting the \nth{2} iteration from a more \enquote{advanced} position on the previous Pareto front.
		We run and average $M=2$ fine-tunings from each of those two initializations for $T=7k$ steps, before applying \LITI.
		Our results indicate that a higher $\eta$ (0.5) performs better in regions of high \KL, whereas a lower $\eta$ (0.3) helps in regions with \KL below 65.
		This suggests that the optimal choice for $\eta$ is compute-dependent; a lower $\eta$ is appropriate if further iterations can explore high \KL regions, whereas a limited compute budget might benefit from a higher $\eta$.
		This resembles the learning rate trade-off in optimization, where lower rates improve results but require more training steps.
		As a final note, we can also use different $\eta$ for the different fine-tunings; notably, we observe that merging all those $M=4$ RLs perform better (though it doubles the compute).
		}
	\label{app:fig:mulr}%
\end{figure*}%
\FloatBarrier
\subsection{Interpolate towards the initialization? or towards the SFT?}
\label{app:expe:ablationinit}
\FloatBarrier
\begin{figure*}[h!]
	\begin{center}
		\begin{subfigure}[b]{0.325\textwidth}%
			\includegraphics[width=\textwidth]{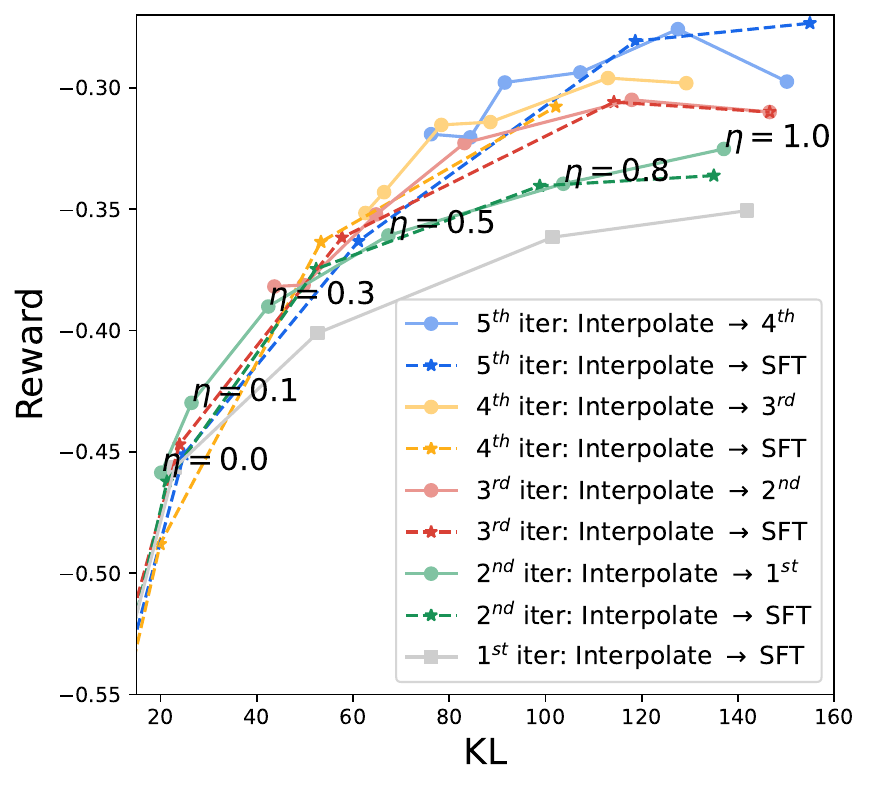}
		\end{subfigure}
	\end{center}%
	\caption{
		\textbf{Experiments ablating the initialization in \LITI.}
		We compare linear interpolating either towards the episode-specific initialization (\ie the $\theta^{\eta}$ selected from previous iteration) or towards the SFT, which was the initialization of the \nth{1} episode.
		The two resulting Pareto fronts are similar.
		However, in our iterative experiments we interpolate towards the episode-specific initialization as it allows maintaining a constant $\eta$ at each \WARP iteration, enabling a smooth progression towards the high \KL regions.}%
	\label{app:fig:tosft_controlvskl}%
\end{figure*}%

\FloatBarrier
\clearpage
\section{Addressing length bias in \WARP}
\label{app:length_penalty}
\begin{figure*}[h!]
	\begin{center}
		\begin{subfigure}[b]{0.325\textwidth}%
			\includegraphics[width=\textwidth]{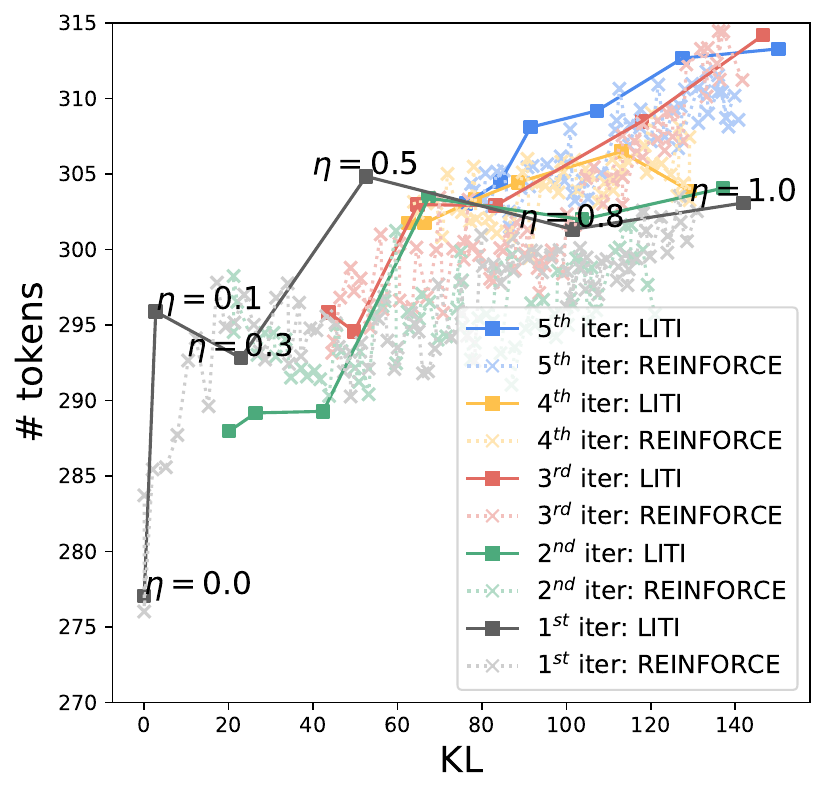}
			\caption{Length bias in iterative \WARP.}%
			\label{app:fig:warpi_numtokensvskl}%
		\end{subfigure}
		\hfill		
		\begin{subfigure}[b]{0.325\textwidth}%
			\includegraphics[width=\textwidth]{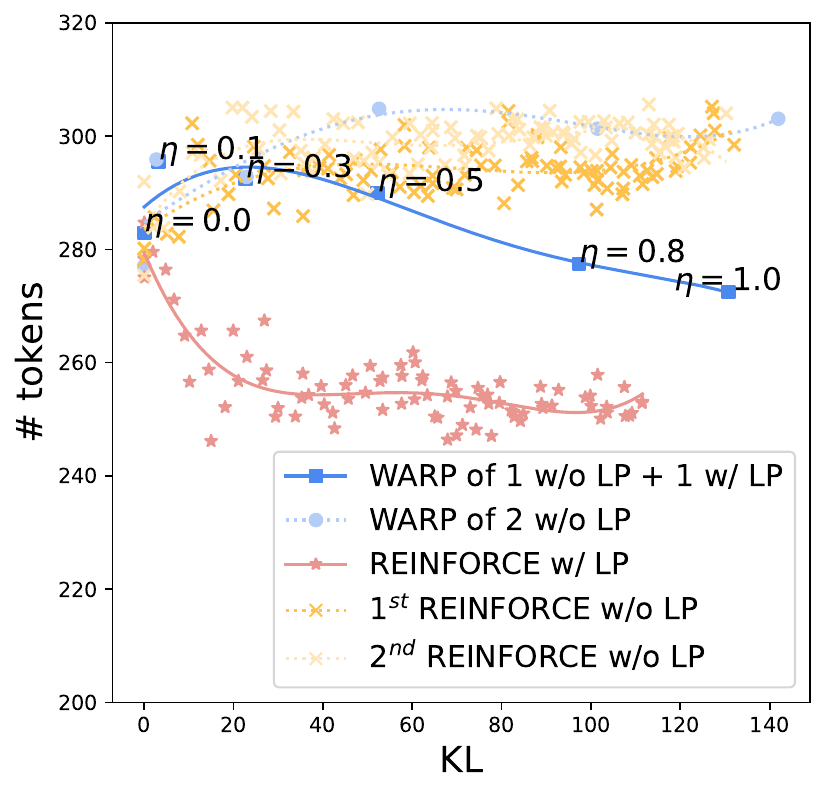}
			\caption{Adding length penalty (LP).}%
			\label{app:fig:diversityLP_numtokensvskl}%
		\end{subfigure}
		\hfill
		\begin{subfigure}[b]{0.325\textwidth}%
			\includegraphics[width=\textwidth]{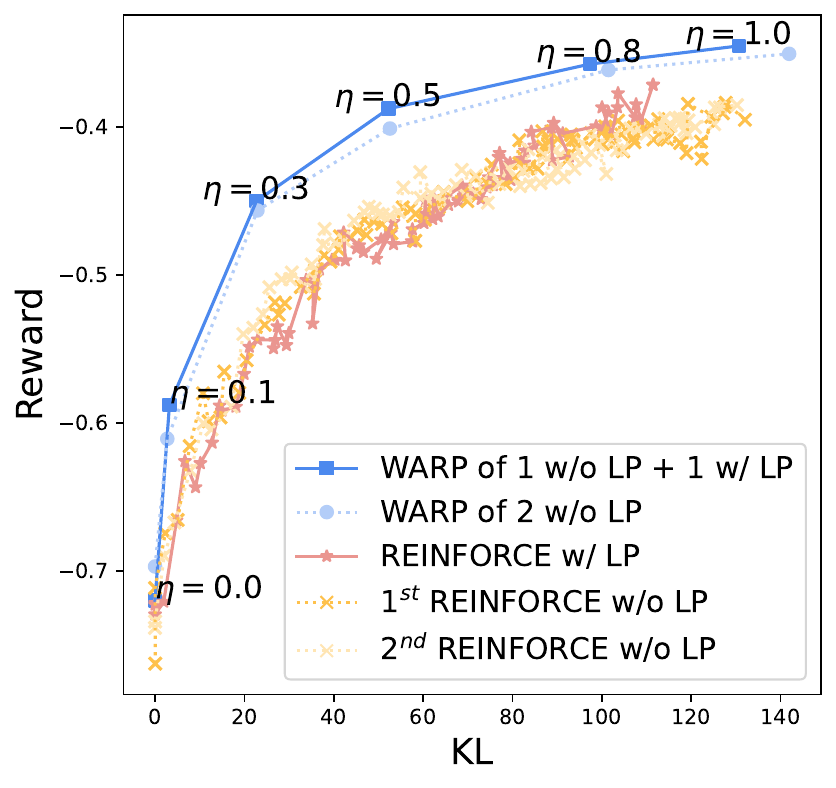}
			\caption{Benefits of diversity.}%
			\label{app:fig:diversityLP_controlvskl}%
		\end{subfigure}	
	\end{center}%
	\caption{\textbf{Addressing length bias in \WARP}.
	\Cref{app:fig:warpi_numtokensvskl} explores how length and \KL change in successive \WARP iterations.
	\Cref{app:fig:diversityLP_numtokensvskl} demonstrates the effectiveness of length penalty (LP) in reducing output length, and how such policies can be merged with others trained without LP.
	Finally, \Cref{app:fig:diversityLP_controlvskl} suggest an additional benefit of merging policies with different objectives, as it improves the \KL-reward Pareto front.
	}%
	\label{app:fig:length_penalty}%
\end{figure*}%

\textbf{Problem: length bias.}
We investigate a potential length bias in \WARP \cite{shen2023loose}. LLMs after \RLHF tend to be unnecessarily verbose because RMs often prefer longer generations to shorter ones, leading to this form of reward hacking. We confirm such a phenomenon in \Cref{app:fig:warpi_numtokensvskl}, where the length of the generated text increases with higher \KL values. This trend is even more pronounced in iterative \WARP, where the \nth{3} iteration generates longer sentences than the \nth{1} iteration at the same \KL.

\textbf{Mitigation strategy: length penalty.}
To mitigate this length bias, we integrate a length penalty (LP) into the reward: $-0.0005 \times \text{len}(y)$, following \cite{singhal2023long}. From the \SFT init, we launch one RL fine-tuning run with this LP, highlighted with red stars in \Cref{app:fig:diversityLP_numtokensvskl}. This LP leads to significantly shorter outputs as training occurs and \KL increases, in contrast to policies trained without LP.

\textbf{\SLERP with different configurations.}
\Cref{app:fig:diversityLP_numtokensvskl} also displays the lengths of generations from a \SLERP merging of two policies, one trained with the LP and the other without.
Critically, merging policies from diverse training configurations not only mitigates the length bias but also improves the Pareto front, as illustrated in \Cref{app:fig:diversityLP_controlvskl}. This improvement is likely due to the increased diversity in weights and predictive mechanisms across policies, which seems beneficial for generalization, as shown in supervised learning \cite{rame2022diwa}.

\textbf{Conclusion.}
Those experiments highlight the possibility to fix the length bias, and also the benefits of merging policies trained with diverse rewards, supporting previous suggestions from~\cite{rame2023rewarded}.

\clearpage

\FloatBarrier
\section{Diversity in predictions}
\label{app:dlp}
\FloatBarrier
\begin{figure*}[h!]
	\begin{center}
		\begin{subfigure}[b]{0.325\textwidth}%
			\includegraphics[width=\textwidth]{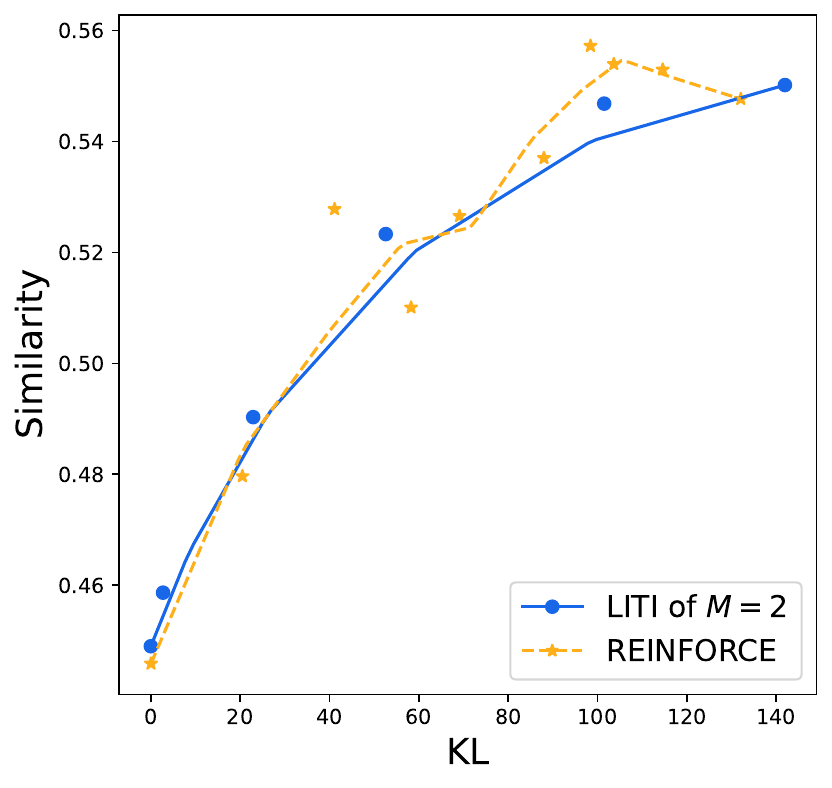}
		\end{subfigure}
	\end{center}%
	\caption{\textbf{Confirming diversity loss in \RLHF}. The $x$-axis is the \KL compared to the \SFT initialization; the $y$-axis is the similarity across two generations from a given policy when decoding with temperature $0.9$.}%
	\label{app:fig:dlp}%
\end{figure*}%
Finally, we investigate the loss in diversity across generations when aligning LLMs, as reported in \cite{kirk2024understanding}.
This could have negative consequences for creative or exploratory tasks, or even lead to policy collapse \cite{Moalla2024NoRN,hamilton2024detecting}.
In \Cref{app:fig:dlp} we plot the BLEURT similarity \cite{sellam-etal-2020-bleurt} across generations, during REINFORCE, or in \LITI (as we interpolate back towards the \SFT initialization).
We observe that \KL is strongly positively correlated with similarity across generations, confirming that \RLHF induces a loss of diversity across generations.
This experiment confirms that optimizing the Pareto optimality between reward and \KL enables to trade-off between alignment and other benefits from pre-training, such as diversity in generations.
\FloatBarrier

\end{document}